\documentclass[journal]{IEEEtran}
\usepackage{cite}
\usepackage{graphicx}
\usepackage{url}
\usepackage{amsmath}
\usepackage[caption = false, font=footnotesize]{subfig}
\usepackage{float}
\usepackage{booktabs} 
\usepackage{amsfonts}

\usepackage{array}
\usepackage{color}
\usepackage{tabularx}
\usepackage{longtable}
\usepackage{tabu}
\usepackage[colorlinks,linkcolor=black]{hyperref}
\usepackage{multirow}
\newcolumntype{L}[1]{>{\raggedright\let\newline\\\arraybackslash\hspace{0pt}}m{#1}}
\newcolumntype{C}[1]{>{\centering\let\newline\\\arraybackslash\hspace{0pt}}m{#1}}
\newcolumntype{R}[1]{>{\raggedleft\let\newline\\\arraybackslash\hspace{0pt}}m{#1}}
\newcommand{\etal}{\textit{et al}.}
\newcommand{\ie}{\textit{i}.\textit{e}.}
\newcommand{\eg}{\textit{e}.\textit{g}.}

\usepackage{blkarray, bigstrut}

\definecolor{grey}{RGB}{130,130,130} 
\definecolor{black}{RGB}{0,0,0} 

\definecolor{red}{RGB}{255,0,0}

\makeatletter
\newcommand*{\rom}[1]{\expandafter\@slowromancap\romannumeral #1@}
\makeatother

\ifCLASSINFOpdf
\else
\fi

\begin{document}
%
\title{Task-Specific Normalization for Continual Learning of Blind Image Quality Models}
%
%
%

\author{Weixia~Zhang,~\IEEEmembership{Member,~IEEE,}
        Kede~Ma,~\IEEEmembership{Senior Member,~IEEE,}
        Guangtao~Zhai,~\IEEEmembership{Senior Member,~IEEE,}
        and~Xiaokang~Yang,~\IEEEmembership{Fellow,~IEEE}
\thanks{Weixia Zhang, Guangtao Zhai, and Xiaokang Yang are with the MoE Key Lab of Artificial Intelligence, AI Institute, Shanghai Jiao Tong University, Shanghai, China (e-mail: zwx8981@sjtu.edu.cn; zhaiguangtao@sjtu.edu.cn; xkyang@sjtu.edu.cn).}
\thanks{Kede Ma is with the Department of Computer Science and  Shenzhen Research Institute, City University of Hong Kong, Kowloon, Hong Kong (e-mail: kede.ma@cityu.edu.hk).}
}

\markboth{IEEE Transactions on Image Processing}%
{Shell \MakeLowercase{\textit{et al.}}: Bare Demo of IEEEtran.cls for Journals}

\maketitle

\begin{abstract}
In this paper, we present a simple yet effective continual learning method for blind image quality assessment (BIQA) with improved quality prediction accuracy, plasticity-stability trade-off, and task-order/-length robustness. The key step in our approach is to freeze all convolution filters of
a pre-trained deep neural network (DNN) for an explicit promise of stability, and learn task-specific normalization parameters for plasticity. We assign each new IQA dataset (\ie, task) a prediction head, and load the corresponding normalization parameters to produce a quality score. The final quality estimate is computed by \textcolor{black}{a weighted summation of predictions from all heads with a lightweight $K$-means gating mechanism}. Extensive experiments on six IQA datasets demonstrate the advantages of the proposed method in comparison to previous training techniques for BIQA.
\end{abstract}

\begin{IEEEkeywords}
Blind image quality assessment, continual learning, task-specific normalization.
\end{IEEEkeywords}

%
\IEEEpeerreviewmaketitle
\section{Introduction}\label{sec:intro}
\begin{figure}
  \centering
  \includegraphics[width=.50\textwidth]{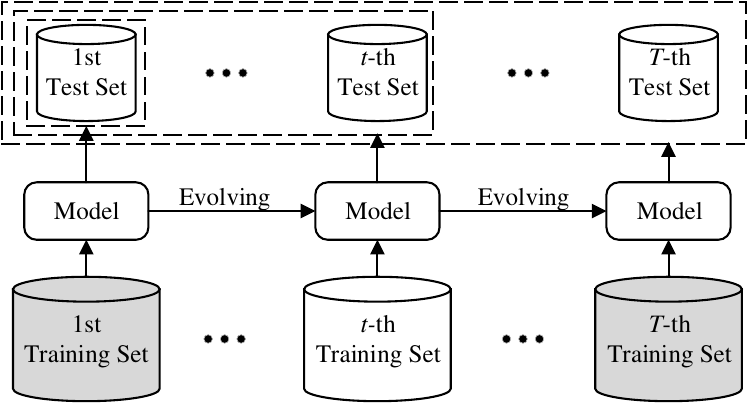}
  \caption{Illustration of continual learning for BIQA. The grey cylinders denote the inaccessibility of previous and future training data. During testing, we use all previous and the current test sets (indicated by dashed rectangles) to evaluate the stability and plasticity of the learned BIQA model.}\label{fig:illustration}
\end{figure}

\IEEEPARstart{T}{here} is an emerging trend to develop image quality assessment (IQA) models~\cite{wang2006modern} and image processing methods in alternation: Better IQA models provide more reliable guidance to the design and optimization of the latter, while new image processing algorithms call for the former to handle novel visual artifacts. This suggests a desirable IQA model to easily adapt to such distortions by continually learning from new data (see Fig.~\ref{fig:illustration}). 

This paper focuses on continual learning of blind IQA (BIQA) models \cite{zhang2023continual,liu2021liqa}, which predict the perceptual quality of a ``distorted'' image without reference to an original undistorted counterpart. Over the past $20$ years, the research in BIQA has shifted from handling distortion-specific~\cite{wang2002no}, single-stage \cite{sheikh2006statistical}, synthetic artifacts to general-purpose~\cite{mittal2012no}, multi-stage \cite{Jayaraman2013Objective}, authentic ones, and from relying on \textcolor{black}{handcrafted} features to purely data-driven approaches \cite{kang2014convolutional}. Existing BIQA models are generally developed and tested using human-rated images from the same dataset, \ie, within the same subpopulation \cite{zhang2023continual}. As such, even the best-performing BIQA methods, \eg, those
 based on deep neural networks (DNNs)  are bound to encounter subpopulation shift when deployed in the real world. 
 
Direct fine-tuning model parameters with new data may result in \textit{catastrophic forgetting}~\cite{mcclelland1995there,mccloskey1989catastrophic} of previously seen data. The dataset combination trick in~\cite{zhang2021uncertainty}  has been proven effective in handling subpopulation shift, but is limited by the computational  scalability and the dataset accessibility. Recently, Zhang~\etal~\cite{zhang2023continual} formulated continual learning for BIQA with five desiderata. They 
also described the first continual learning method of training BIQA models based on a technique called learning with forgetting (LwF) \cite{li2018learning}. Like many continual learning methods (for classification), LwF adds a form of regularization~\cite{de2021continual} to mildly adjust model parameters for new tasks while respecting old tasks. Nevertheless,  this type of regularization-based methods have two limitations. First, it is practically difficult to set the trade-off parameter for stability (\ie, the ability to 
consolidate acquired knowledge from old tasks) and plasticity (\ie, the ability to learn new knowledge from the current task). Second, the performance is usually sensitive to the order and the length of the task sequence \cite{yoon2020scalable,zhang2023continual}.

In this paper, we describe a simple yet effective continual learning method for BIQA based on parameter decomposition. We start with a pre-trained DNN as the feature extractor. We freeze all convolution filters, and share them \textcolor{black}{along with all parameter-free nonlinear activation and pooling layers} across tasks during the entire continual learning process. We append a prediction head, implemented by a fully connected (FC) layer, when learning a new task.
We allow the parameters of batch normalization (BN) \cite{ioffe2015batch} following each convolution to be specifically learned for each task. Through this \textit{task-specific} normalization, a better plasticity-stability trade-off can be made with a negligible increase in model size.  During inference, we load each group of BN parameters to produce a quality estimate using the corresponding prediction head. The final quality score is computed by \textcolor{black}{a weighted summation of predictions from all heads with a lightweight $K$-means gating (KG) mechanism.}

In summary, our contributions are threefold.

\begin{itemize}\setlength{\itemsep}{2pt}
    \item We propose \textcolor{black}{a new continual learning method for BIQA. The resulting method, which we name TSN-IQA, integrates new knowledge into BN parameters without catastrophic forgetting of acquired knowledge.}
    \textcolor{black}{\item We design a lightweight KG module that only requires learning a  set of distortion-aware BN parameters (instead of relying on an extra DNN~\cite{zhang2023continual}) to compute the weightings of prediction heads during inference.} 
    \item We perform extensive experiments to demonstrate the advantages of our method in terms of quality prediction accuracy,  plasticity-stability trade-off, and task-order/-length robustness.

\end{itemize}

\section{Related Work}\label{sec:related}
In this section, we give an overview of recent progress in BIQA. We then review representative continual learning methods for classification, and discuss normalization techniques in the broader context of deep learning.
\subsection{BIQA Models}\label{subsec:biqa_methods}
Many early  BIQA methods are based on hand-engineered natural scene statistics (NSS) in spatial~\cite{mittal2012no, mittal2013making}, transformed~\cite{moorthy2011blind}, or both domains~\cite{ghadiyaram2017perceptual}. In recent years, deep learning began to show its promise in the field of BIQA. Patchwise training~\cite{bosse2016deep,kang2014convolutional}, transfer learning~\cite{zeng2018blind}, and quality-aware pre-training~\cite{liu2017rankiqa, Ma2018End, ma2019blind} were proposed to compensate for the lack of human-rated data. Of particular interest is the introduction of IQA datasets with realistic distortions~\cite{ciancio2011no, ghadiyaram2016massive, hosu2020koniq}, which excites a series of BIQA models to address the synthetic-to-real generalization. Zhang~\etal~\cite{zhang2020blind} assembled two network branches to account for synthetic and realistic distortions separately. Su~\etal~\cite{9156687} investigated content-aware convolution for robust BIQA, while Zhu~\etal~\cite{zhu2020metaiqa} aimed to learn more transferable quality-aware representations by meta-learning. Zhang~\etal~\cite{zhang2021uncertainty} proposed a dataset combination strategy to train BIQA models on multiple IQA datasets. They later \textcolor{black}{formulated the continual learning setting for BIQA, and} introduced a method \textcolor{black}{that combines} LwF~\cite{li2018learning} with a \textcolor{black}{\textit{K}-means gating (KG) module}~\cite{zhang2023continual}. Concurrently, Liu~\etal~\cite{liu2021liqa} proposed a continual learning method for BIQA based on a replay strategy. \textcolor{black}{Ma~\etal~\cite{ma2021remember} relied on model pruning techniques to enable continual learning of BIQA methods.} 

\textcolor{black}{In this paper, we follow the setting in~\cite{zhang2023continual}, and propose a new continual learning method for BIQA with significantly improved performance in several aspects.}

\subsection{Continual Learning for  Classification}\label{subsec:cl}
While humans rarely forget previously learned knowledge catastrophically, machine learning models such as DNNs tend to do so when learning new concepts~\cite{french1999catastrophic, mccloskey1989catastrophic}. Enforcing regularization is a common practice to mitigate the catastrophic forgetting problem in continual learning. For example, Li and Hoiem~\cite{li2018learning} proposed LwF, which leverages model predictions of previous tasks as pseudo labels. Elastic weight consolidation (EWC)~\cite{kirkpatrick2017overcoming}, variational continual learning (VCL)~\cite{nguyen2018variational}, synaptic intelligence (SI)~\cite{zenke2017continual}, and memory-aware synapses (MAS)~\cite{aljundi2018memory} work similarly by identifying and penalizing changes to important parameters of previous tasks. From this perspective, parameter decomposition~\cite{de2021continual} can be seen as a form of hard regularization, disentangling model parameters into task-agnostic and task-specific groups. This may be done by either masking learned parameters of previous tasks~\cite{fernando2017pathnet, mallya2018piggyback, mallya2018packnet} or growing new branches to accommodate new tasks~\cite{rusu2016progressive}. For example, Yoon~\etal~\cite{yoon2020scalable} proposed additive parameter decomposition to achieve task-order robustness. Singh~\etal \cite{singh2020calibrating} calibrated the convolution responses  of a continually trained DNN with a few parameters for new tasks. In this paper, we take a similar but much simpler parameter decomposition approach to achieve accurate and robust continual learning for BIQA.

\subsection{Normalization in Deep Learning}\label{subsec:normalization}
There is increasing evidence that normalization is a canonical neural computation throughout the
visual system, and in many other sensory modalities and brain regions \cite{carandini2012normalization}. As biologically inspired, deep learning also incorporates different instantiations of  normalization for various purposes, such as accelerating model training \cite{ioffe2015batch} and improving model generalization~\cite{huang2020normalization}. BN is a \textcolor{black}{popular} technique to improve the training efficiency of DNNs, in which the convolution responses are divided by the standard deviation (std) of a pool of responses along the batch (and spatial) dimensions. Xie~\etal~\cite{xie2020adversarial} learned separate BN layers to harness adversarial examples, which improves image recognition models. Li~\etal~\cite{li2017revisiting} proposed adaptive BN for domain adaptation, assuming that domain-invariant and domain-specific computations are learned by the convolution filters and the BN layers, respectively. Chang~\etal~\cite{chang2019domain} specialized BN layers using a two-stage algorithm for unsupervised domain adaptation. Dumoulin~\etal~\cite{dumoulin2017learned} relied on conditional instance normalization~\cite{ulyanov2016instance} to synthesize the artistic styles of diverse paintings. Zhang~\etal~\cite{zhang2020passport} presented a passport normalization for deep model intellectual property protection. 
In this paper,  we introduce task-specific BN to accomplish continual learning of DNN-based BIQA models.


\section{Proposed Method}\label{sec:cnique}
In this section, we first revisit the  formulation of continual learning for BIQA in~\cite{zhang2023continual}, and then elaborate the training and inference procedures of the proposed \textcolor{black}{TSN-IQA}. \textcolor{black}{To facilitate mathematical comprehension, we summarize a list of variables in Table~\ref{tab:variable_list}.}
\subsection{Problem Formulation}
When training on the $t$-th dataset $\mathcal{D}_t$, \ie, the $t$-th task, a BIQA model $f_w$, parameterized by a vector $w$, has no direct access to previous training images in $\{\mathcal{D}_k\}_{k=1}^{t-1}$, leading to the following objective:
\begin{align}\label{eq:tra}
\mathcal{L}(\mathcal{D}_t;w) =\frac{1}{\vert\mathcal{D}_t\vert}\sum_{(x,\textcolor{black}{\mu_x})\in \mathcal{D}_t}\ell(f_w(x),\textcolor{black}{\mu_x}) +\lambda \textcolor{black}{\Omega}(w),
\end{align}
where $x$ and \textcolor{black}{$\mu_x$} denote the ``distorted'' image and the corresponding mean opinion score (MOS), respectively. $\ell(\cdot)$ is a quantitative measure of quality prediction performance, and \textcolor{black}{$\Omega(\cdot)$} is an optional regularizer. A good BIQA model under this setting should adapt well to new tasks, and meanwhile endeavor to mitigate catastrophic forgetting of old tasks as measured by
\begin{align}\label{eq:def}
\sum_{k=1}^{t}\mathcal{L}(\mathcal{V}_k;w) = \sum_{k=1}^{t}\left(\frac{1}{\vert\mathcal{V}_k\vert}\sum_{(x,\textcolor{black}{\mu_x})\in \mathcal{V}_k}\ell(f_w(x),\textcolor{black}{\mu_x}) \right),
\end{align}
where $\mathcal{V}_k$ denotes the test set for the $k$-th task. Five desiderata are suggested in~\cite{zhang2023continual} to make continual learning for BIQA feasible and nontrivial: 1) common perceptual scale, 2) robust to subpopulation shift, 3) limited access to previous data, 4) no test-time oracle, and 5) bounded memory footprint.

\begin{table}[tbp]
  \centering
  {\color{black}\caption{List of variables}}\label{tab:variable_list}
  \begin{tabular}{l|l}
      \toprule
    \color{black}{Notation} & \color{black}{Description}\\
    \hline
    \color{black}{$(x, y)$}  & \color{black}{an image pair}\\
    \color{black}{$(\mu_x$, $\mu_y)$}  & \color{black}{MOSs of $x$ and $y$}\\
    \color{black}{$\mathcal{D}_t$}  & \color{black}{the $t$-th dataset}\\
    \color{black}{$\mathcal{P}_t$}  & \color{black}{the $t$-th paired dataset}\\
    \color{black}{$N_{t}$}  & \color{black}{\# of image pairs in the $t$-th paired dataset}\\
    \color{black}{$f_\phi$}  & \color{black}{a DNN parameterized by a vector $\phi$}\\
    \color{black}{$h_{\psi_t}$}  & \color{black}{the $t$-th prediction head parameterized by $\psi_t$}\\
    \color{black}{$r(x, y)$}  & \color{black}{the binary quality label of $(x,y)$}\\
    \color{black}{$\hat{p}_t(x,y)$}  & \color{black}{the predicted probability of $(x,y)$ for the $t$-th task}\\
    \color{black}{$(\mu_t, \sigma_t, \gamma_t, \beta_t)$}  & \color{black}{the 4-tuple BN parameters for the $t$-th task}\\
    \color{black}{$c_{kst}$}  & \color{black}{the $k$-th centroid at the $s$-th stage for the $t$-th task }\\
    \color{black}{$d_{st}(x)$}  & \color{black}{the perceptual relevance of $x$ to the $t$-th task}\\
    \color{black}{$a_{st}(x)$}  & \color{black}{the weighting of $x$ for the $t$-th prediction head}\\
    \color{black}{$\hat{q}(x)$}  & \color{black}{the predicted quality score of $x$}\\
     \bottomrule
  \end{tabular}
\end{table}

\subsection{Model Estimation}\label{subsec:estimation}
 Inspired by UNIQUE~\cite{zhang2021uncertainty}, we exploit relative quality information to learn a common perceptual scale for all tasks.  Specifically, given an image pair $(x, y)$,  we compute a binary label:
\begin{align}\label{eq:bgt}
     r(x,y) = 
\begin{cases} 
 1 & \mbox{if } \mu_{x}\ge \mu_{y} \\
      0 & \mbox{otherwise} \end{cases}.
\end{align}
 Careful readers may find that we do not infer a continuous value $p(x,y)$, which denotes the probability of  $x$ perceived better than $y$ based on the Thurstone's model \cite{thurstone1927law} or the Bradley-Terry model \cite{bradley1952rank} as typically done in previous work \cite{ma2019blind,zhang2021uncertainty}. This is because the computed probability may vary with the precision of the subjective testing methodology. For example, if $x$ is  marginally better than $y$ and a precise subjective method such as the two-alternative forced choice (2AFC) is adopted, $p(x,y)$ can be close to one. By contrast, if a less precise subjective method such as the single stimulus continuous quality rating is used, $p(x,y)$ may only be slightly larger than $0.5$. Compared to $p(x,y)$,  we also empirically observe that $r(x,y)$ leads to faster convergence and improved accuracy results. When learning the $t$-th task, we transform $\mathcal{D}_t =\{x^{(i)}_t, \mu^{(i)}_t\}_{i=1}^{\vert\mathcal{D}_t\vert}$ to $\mathcal{P}_t=\{(x^{(i)}_t,y^{(i)}_t), r^{(i)}_t\}_{i=1}^{N_t}$, where $N_t \le \binom{\vert\mathcal{D}_t\vert}{2}$.

\begin{figure}
  \centering
  \includegraphics[width=.4\textwidth]{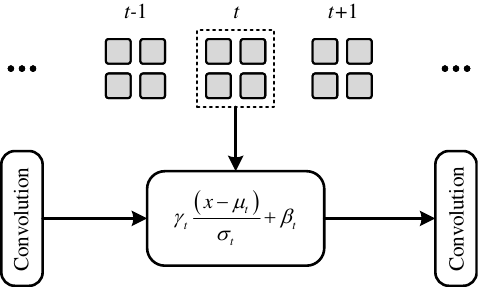}
  \caption{Illustration of task-specific BN. The parameters of all convolutions are frozen and shared across all tasks. A group of BN parameters is customized for each task.}\label{fig:cbn}
\end{figure}

Our BIQA model consists of a feature extractor implemented by a DNN, $f_\phi(\cdot)$, producing a fixed-length image representation independent of input resolution. For the $t$-th task, we append a prediction head implemented by an FC layer, $h_{\psi_t}(\cdot)$, outputting a corresponding quality score. Under the Thurstone's Case V model \cite{thurstone1927law}, we estimate the probability that $x$ is of higher quality than $y$ by
\begin{align}\label{eq:difference2}
\hat{p}_t(x,y)= \Phi\left(\frac{h_{\psi_t}(f_{\phi}(x)) - h_{\psi_{t}}(f_{\phi}(y))}{\sqrt{2}}\right),
\end{align}
where the quality prediction variance is fixed to one. We measure the statistical distance between the ground-truth labels and predicted probabilities using the fidelity loss~\cite{tsai2007frank} due to its favorable optimization behaviors~\cite{zhang2021uncertainty}:
\begin{align}\label{eq:fidelity}
\ell(x, y;\phi, \psi_t)
= 1& - \sqrt{r(x,y)\hat{p}_t(x,y)} \nonumber \\ &-\sqrt{(1-r(x,y))(1-\hat{p}_t(x,y))}.
\end{align}

To make a better trade-off between plasticity and stability while keeping a bounded model size, our  BIQA method chooses to maximally share computation across tasks, and customize a tiny fraction of parameters to account for the incremental difference introduced by new tasks. In particular, our feature extractor is composed of several stages of convolution, BN, halfwave-rectification (\ie, ReLU nonlinearity), and max-pooling. We freeze all pre-trained convolution parameters during model development, and learn a group of $4$-tuple BN parameters for the $t$-th task
 \begin{align}\label{eq:bn}
z_{\mathrm{BN}} = \gamma_t\left(\frac{z - \mu_t}{\sigma_t}\right) + \beta_t,
\end{align}
where $\mu_t$ and $\sigma_t$ are the mean and the std estimated by the exponentially decaying moving average over mini-batches. $\gamma_t$ and $\beta_t$ are the learnable scaling and shift parameters (see also Fig.~\ref{fig:cbn}). After training on a $T$-length task sequence, we obtain $T$ groups of task-specific BN parameters. 

\subsection{Model Inference}\label{subsec:inference}
During inference, \textcolor{black}{we successively load each of $T$ groups of BN parameters along with the corresponding FC layer to compute $T$ quality scores.} Due to the unavailability of the task oracle, we \textcolor{black}{rely on an improved KG module~\cite{zhang2023continual} with a lightweight design goal}, \textcolor{black}{which} is made possible by the proposed parameter decomposition scheme.
\textcolor{black}{Unlike~\cite{zhang2023continual}, we only train a set of task-agnostic BN parameters on a large-scale image set with various synthetic distortions~\cite{zhang2020blind} for distortion-aware weighting computation, while keeping all convolution filters intact.} \textcolor{black}{Since  the original BN parameters of the pre-trained feature extractor are not necessary, our gating mechanism introduces essentially no extra parameters, and adheres to the desideratum of bounded memory footprint}. 

\begin{figure}
  \centering
  \includegraphics[width=.49\textwidth]{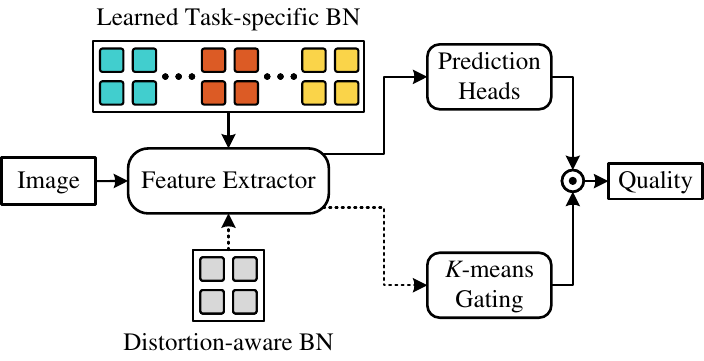}
  \caption{Overview of the inference process. \textcolor{black}{The solid line indicates the pipeline of loading task-specific BN parameters learned for different tasks (denoted by different colors) and making the quality prediction using the corresponding prediction heads. The dotted line indicates the process of loading  distortion-aware BN parameters and computing  weightings for all prediction heads with the KG module.}}\label{fig:inference}
\end{figure}

We present the overview of the inference process in Fig.~\ref{fig:inference}. During learning the $t$-th task, we \textcolor{black}{load the distortion-aware BN parameters to} the pre-trained $f_\phi$ to compute globally pooled convolution responses of image $x$ at the $s$-th stage, $\bar{f}_{\phi_s}(x)$.
Given $S$-stage convolutions, we obtain a feature summary of $\mathcal{D}_t$:  $\{\bar{f}_{\phi_1}(x_t^{(i)}),\ldots, \bar{f}_{\phi_S}(x_t^{(i)})\}_{i=1}^{\vert\mathcal{D}_t\vert}$. We then apply $K$-means~\cite{lloyd1982least} (for each stage of convolution responses) to compute $S$ groups of $K$ centroids $\{\{c_{kst}\}_{k=1}^{K}\}_{s=1}^{S}$. 

We measure the perceptual relevance of $x$ to $\mathcal{D}_t$ by computing the minimal Euclidean distances between $\bar{f}_{\phi_s}(x)$ and $\{c_{kst}\}_{k=1}^{K}$:
\begin{align}\label{min_distance}
d_{st}(x) = \min_{k}\|\bar{f}_{\phi_s}(x) - c_{kst}\|_2\,.
\end{align}
We pass $\{d_{st}(x)\}_{t=1}^T$ to a softmin function to compute the \textcolor{black}{weightings} at the $s$-th stage for the $t$-th prediction head:
\begin{align}\label{eq:softmin}
a_{st}(x) = \frac{{\rm exp}(-{\tau}d_{st}(x))}{\sum_{t=1}^{T}{\rm exp}(-{\tau}d_{st}(x))},
\end{align}
where $\tau \geq 0$ is a  parameter to control the smoothness of the softmin function. We further average the weightings across stages to obtain
\begin{align}
a_{t}(x) = \frac{1}{S}\sum_s a_{st}(x).
\end{align}
We last compute the overall quality score by the inner product between the weighting and quality prediction vectors:
\begin{align}\label{eq:quality}
\hat{q}(x) = \sum_{t=1}^{T}a_{t}(x) h_{\psi_{t}}(f_\phi(x)).
\end{align}

\section{Experiments}\label{sec:exp}
In this section, we first describe the experimental setups for continual learning of BIQA models, and then compare the proposed \textcolor{black}{TSN-IQA}  against previous training techniques,  supplemented by abundant ablation studies. The source code is made publicly available at \url{https://github.com/zwx8981/TSN-IQA} for reproducible research.

\begin{table*}[htbp]
  \centering
  \small
  \caption{Summary of IQA datasets used in our experiments. CLIVE stands for the LIVE Challenge Database. TID2013~\cite{ponomarenko2013color}, SPAQ~\cite{fang2020perceptual}, \textcolor{black}{and AGIQA-3K~\cite{li2023agiqa}} are used for cross-database evaluation. SS: Single stimulus.  DS: Double stimulus. MS: Multiple stimulus. CQR: Continuous quality rating. ACR: Absolute category rating. CS: Crowdsourcing. PC: Paired Comparison}\label{tab:database}
  \begin{tabular}{l|ccccccc}
      \toprule
        {Dataset} & \# of Images & \# of Training Pairs & \# of Test Images & Scenario & \# of Types & Testing Methodology& Year \\
     \hline
        LIVE~\cite{sheikh2006statistical} & 779 & 7,780 & 163 & Synthetic & 5 & SS-CQR &2006 \\
        CSIQ~\cite{larson2010most} & 866 & 8,786 & 173 & Synthetic & 6 & MS-CQR & 2010\\
        BID~\cite{ciancio2011no} & 586 & 11,204 & 117 & Realistic & N.A. & SS-CQR & 2011\\
        CLIVE~\cite{ghadiyaram2016massive} & 1,162 & 24,604 & 232 & Realistic & N.A. & SS-CQR-CS  & 2016 \\
        KonIQ-10K~\cite{hosu2020koniq} & 10,073 & 139,274 & 2,015 & Realistic & N.A. & SS-ACR-CS & 2018\\
        KADID-10K~\cite{lin2019kadid} & 10,125 & 140,071 & 2,000 & Synthetic & 25 & DS-ACR-CS & 2019\\
        \hline
        TID2013~\cite{ponomarenko2013color} & 3,000 & N.A. & 3,000 & Synthetic & 25 & DS-PC & 2013\\
        SPAQ~\cite{fang2020perceptual} & 11,125 &  N.A. & 11,125 & Realistic &  N.A. & SS-CQR & 2020\\
        \textcolor{black}{AGIQA-3K~\cite{li2023agiqa}} & \textcolor{black}{2,982} &  \textcolor{black}{N.A.} & \textcolor{black}{2,982} & \textcolor{black}{Generated} &  \textcolor{black}{N.A.} & \textcolor{black}{SS-CQR} & \textcolor{black}{2023}\\
     \bottomrule
  \end{tabular}
\end{table*}

\begin{table*}[t]
  \centering
  \caption{Performance comparison in terms of $\mathrm{mSRCC}$, $\mathrm{mPI}$, $\mathrm{mSI}$, and $\mathrm{mPSI}$. Task-aware and task-agnostic evaluation settings correspond to quality prediction with and without the task oracle, respectively. All methods  are trained in chronological order}\label{tab:mpsr}
  \begin{tabular}{c|l|cccc|cccc}
      \toprule
    \multirow{2}{*}{Setting}& \textcolor{black}{Backbone} & \multicolumn{4}{c|}{\textcolor{black}{ResNet-18}} & \multicolumn{4}{c}{\textcolor{black}{Two-Stream DNN}} \\ 
    \cline{2-10}
      & {Method} & $\mathrm{mSRCC}$ & $\mathrm{mPI}$ & $\mathrm{mSI}$ & $\mathrm{mPSI}$ & \textcolor{black}{$\mathrm{mSRCC}$} & \textcolor{black}{$\mathrm{mPI}$} & \textcolor{black}{$\mathrm{mSI}$} & \textcolor{black}{$\mathrm{mPSI}$}\\
    \hline
     \multirow{4}{*}{{Task-Aware}} & SI-O & 0.786 & \textbf{0.858} & 0.886 & 0.872 & \textcolor{black}{0.834} & \textcolor{black}{\textbf{0.881}} & \textcolor{black}{0.947} & \textcolor{black}{0.914}\\
     & MAS-O & 0.779 & 0.853 & 0.882 & 0.868 & \textcolor{black}{0.835} & \textcolor{black}{0.874} & \textcolor{black}{0.955} & \textcolor{black}{0.915}\\
     & LwF-O & 0.804 & 0.841 & 0.970 & 0.906 & \textcolor{black}{0.849} & \textcolor{black}{0.877} & \textcolor{black}{0.986} & \textcolor{black}{\textbf{0.931}}\\
     & \textcolor{black}{TSN-IQA-O} & \textbf{0.855} & 0.855 & \textbf{1.000} & \textbf{0.928} & \textcolor{black}{\textbf{0.862}} & \textcolor{black}{0.862} & \textcolor{black}{\textbf{1.000}} & \textcolor{black}{\textbf{0.931}}\\
    \hline
     \multirow{8}{*}{Task-Agnostic}& SL & 0.666 & 0.851 & 0.751 & 0.801 & \textcolor{black}{0.677} & \textcolor{black}{0.876} & \textcolor{black}{0.805} & \textcolor{black}{0.840}\\        
    & SI & 0.762 & \textbf{0.858} & 0.876 & 0.867 & \textcolor{black}{0.732} & \textcolor{black}{\textbf{0.881}} & \textcolor{black}{0.822} & \textcolor{black}{0.852}\\
     & SI-KG & 0.778 & 0.856 & 0.877 & 0.867 & \textcolor{black}{0.785} & \textcolor{black}{0.877} & \textcolor{black}{0.932} & \textcolor{black}{0.900}\\  
     & MAS & 0.717 & 0.853 & 0.862 & 0.858 & \textcolor{black}{0.617} & \textcolor{black}{0.874} & \textcolor{black}{0.797} & \textcolor{black}{0.836}\\
     & MAS-KG & 0.769 & 0.854 & 0.872 & 0.863 & \textcolor{black}{0.780} & \textcolor{black}{0.862} & \textcolor{black}{0.943} & \textcolor{black}{0.903}\\
     & LwF & 0.691 & 0.841 & 0.890 & 0.866 & \textcolor{black}{0.669} & \textcolor{black}{0.877} & \textcolor{black}{0.833} & \textcolor{black}{0.855}\\
     & LwF-KG & 0.801 & 0.837 & 0.963 & 0.900 & \textcolor{black}{0.815} & \textcolor{black}{0.856} & \textcolor{black}{0.980} & \textcolor{black}{0.918}\\
     &\textcolor{black}{TSN-IQA}  & \textbf{0.846} & 0.853 & \textbf{0.979} & \textbf{0.916} & \textcolor{black}{\textbf{0.846}} & \textcolor{black}{0.860} & \textcolor{black}{\textbf{0.985}} & \textcolor{black}{\textbf{0.923}}\\
     \bottomrule
  \end{tabular}
\end{table*}

\begin{table*}[htbp]
  \centering
  \caption{Performance comparison in terms of SRCC. Best results in each section are highlighted in bold, and results of future tasks are marked in grey.  \textcolor{black}{All methods employ a variant of ResNet-18 as the backbone network}}\label{tab:srcc_results}
  \begin{tabular}{c|l|cccccc}
      \toprule
      {Dataset} & {Method} & LIVE~\cite{sheikh2006statistical} & CSIQ~\cite{larson2010most} & BID~\cite{ciancio2011no} & CLIVE~\cite{ghadiyaram2016massive} & KonIQ-10K~\cite{hosu2020koniq} & KADID-10K~\cite{lin2019kadid}\\
     \midrule
        All & JL & 0.969 & 0.815 & 0.842 & 0.827 & 0.856 & 0.896\\
        \hline
        \multirow{8}{*}{LIVE} & SL & 0.927 & {\color{grey} 0.645} & {\color{grey} 0.726} & {\color{grey} 0.407} & {\color{grey} 0.645} & {\color{grey} 0.556}\\
        & LwF & 0.927 & {\color{grey} 0.645} &  {\color{grey} 0.726} & {\color{grey} 0.407} & {\color{grey} 0.645} & {\color{grey} 0.556}\\
        & \textcolor{black}{LwF-KG} & 0.927 & {\color{grey} 0.645} & {\color{grey} 0.726} & {\color{grey} 0.407} & {\color{grey} 0.645} & {\color{grey} 0.556}\\
        & SI  & 0.927 & {\color{grey} 0.645} &  {\color{grey} 0.726} & {\color{grey} 0.407} & {\color{grey} 0.645} & {\color{grey} 0.556}\\
        & \textcolor{black}{SI-KG} & 0.927 &  {\color{grey} 0.645} & {\color{grey} 0.726} & {\color{grey} 0.407} & {\color{grey} 0.645} & {\color{grey} 0.556}\\
        & MAS & 0.927 & {\color{grey} 0.645} &  {\color{grey} 0.726} & {\color{grey} 0.407} & {\color{grey} 0.645} & {\color{grey} 0.556}\\
        & \textcolor{black}{MAS-KG}  & 0.927 & {\color{grey} 0.645} & {\color{grey} 0.726} & {\color{grey} 0.407} & {\color{grey} 0.645} & {\color{grey} 0.556}\\
        & \textcolor{black}{TSN-IQA}  & \textcolor{black}{\textbf{0.956}} & {\color{grey} 0.677} & {\color{grey} 0.645} & {\color{grey} 0.465} & {\color{grey} 0.680} & {\color{grey} 0.504}\\
        \hline
        \multirow{8}{*}{CSIQ} & SL & 0.903 & 0.846 & {\color{grey} 0.656} & {\color{grey} 0.426} & {\color{grey} 0.628} & {\color{grey} 0.552}\\
        & LwF & \textbf{0.954} & 0.805 & {\color{grey} 0.692} & {\color{grey} 0.440} & {\color{grey} 0.684} & {\color{grey} 0.581}\\
        & \textcolor{black}{LwF-KG} & 0.923 & 0.815 & {\color{grey} 0.725} & {\color{grey} 0.472} & {\color{grey} 0.680} & {\color{grey} 0.530}\\
        & SI & 0.953 & \textbf{0.880} & {\color{grey} 0.596} & {\color{grey} 0.415} & {\color{grey} 0.646} & {\color{grey} 0.575}\\
        & \textcolor{black}{SI-KG} & 0.940 & 0.874 & {\color{grey} 0.596} & {\color{grey} 0.414} & {\color{grey} 0.643} & {\color{grey} 0.554}\\
        & MAS & 0.948 & 0.874 & {\color{grey} 0.617} & {\color{grey} 0.412} & {\color{grey} 0.659} & {\color{grey} 0.581}\\
        & \textcolor{black}{MAS-KG}& 0.935 & 0.870 & {\color{grey} 0.617} & {\color{grey} 0.412} & {\color{grey} 0.658} & {\color{grey} 0.566}\\
        & \textcolor{black}{TSN-IQA}  & \textcolor{black}{{0.950}} & \textcolor{black}{{0.850}} & {\color{grey} 0.672} & {\color{grey} 0.477} & {\color{grey} 0.696} & {\color{grey} 0.522}\\
        \hline
        \multirow{8}{*}{BID} & SL & 0.716 & 0.569 & 0.766 & {\color{grey} 0.660} & {\color{grey} 0.645} & {\color{grey} 0.374}\\
        & LwF & 0.938 & 0.792 & 0.782 & {\color{grey} 0.572} & {\color{grey} 0.729} & {\color{grey} 0.547}\\
        & \textcolor{black}{LwF-KG} & 0.920 & 0.796 & 0.789 & {\color{grey} 0.511} & {\color{grey} 0.718} & {\color{grey} 0.528}\\
        & SI & 0.886 & 0.835 & \textbf{0.814} & {\color{grey} 0.562} & {\color{grey} 0.718} & {\color{grey} 0.571}\\
        & \textcolor{black}{SI-KG} & 0.885 & 0.839 & 0.813 & {\color{grey} 0.536} & {\color{grey} 0.717} & {\color{grey} 0.570}\\
        & MAS & 0.907 & 0.826 & 0.785 & {\color{grey} 0.556} & {\color{grey} 0.682} & {\color{grey} 0.595}\\
        & \textcolor{black}{MAS-KG} & 0.841 & 0.833 & 0.785 & {\color{grey} 0.541} & {\color{grey} 0.685} & {\color{grey} 0.594}\\
        & \textcolor{black}{TSN-IQA}  & \textcolor{black}{\textbf{0.952}} & \textcolor{black}{\textbf{0.855}} & \textcolor{black}{{0.812}} & {\color{grey} 0.720} & {\color{grey} 0.701} & {\color{grey} 0.494}\\
        \hline
        \multirow{8}{*}{CLIVE} & SL & 0.448 & 0.472 & \textbf{0.839} & \textbf{0.836} & {\color{grey} 0.756} & {\color{grey} 0.306}\\
        & LwF & 0.814 & 0.646 & 0.820 & 0.802 & {\color{grey} 0.774} & {\color{grey} 0.522}\\
        & \textcolor{black}{LwF-KG} & 0.910 & 0.776 & 0.785 & 0.776 & {\color{grey} 0.756} & {\color{grey} 0.552}\\
        & SI & 0.925 & 0.792 & 0.804 & 0.809 & {\color{grey} 0.804} & {\color{grey} 0.536}\\
        & \textcolor{black}{SI-KG} & 0.929 & 0.800 & 0.807 & 0.811 & {\color{grey} 0.803} & {\color{grey} 0.527}\\
        & MAS & 0.911 & 0.759 & 0.818 & 0.821 & {\color{grey} 0.776} & {\color{grey} 0.497}\\
        & \textcolor{black}{MAS-KG} & 0.928 & 0.770 & 0.828 & 0.821 & {\color{grey} 0.774} & {\color{grey} 0.496}\\
        & \textcolor{black}{TSN-IQA}  & \textcolor{black}{\textbf{0.953}} & \textcolor{black}{\textbf{0.837}} & \textcolor{black}{0.833} & \textcolor{black}{0.799} & {\color{grey} 0.730} & {\color{grey} 0.502}\\
        \hline
        \multirow{8}{*}{KonIQ-10K} & SL & 0.785 & 0.708 & 0.768 & 0.723 & \textbf{0.895} & {\color{grey} 0.593}\\
        & LwF & 0.868 & 0.705 & 0.768 & 0.725 & 0.889 & {\color{grey} 0.598}\\
        & \textcolor{black}{LwF-KG} & 0.913 & 0.741 & 0.800 & 0.779 & 0.870 & {\color{grey} 0.617}\\
        & SI & 0.850 & 0.708 & 0.781 & 0.713 & 0.888 & {\color{grey} 0.608}\\
        & \textcolor{black}{SI-KG} & 0.884 & 0.693 & 0.793 & 0.739 & 0.885 & {\color{grey} 0.611}\\
        & MAS & 0.901 & 0.712 & 0.775 & 0.693 & 0.883 & {\color{grey} 0.613}\\
        & \textcolor{black}{MAS-KG} & 0.936 & 0.716 & 0.797 & 0.737 & 0.880 & {\color{grey} 0.620}\\
        & \textcolor{black}{TSN-IQA}  & \textcolor{black}{\textbf{0.959}} & \textcolor{black}{\textbf{0.813}} & \textcolor{black}{\textbf{0.823}} & \textcolor{black}{\textbf{0.796}} & \textcolor{black}{0.869} & {\color{grey} 0.568}\\
        \hline
        \multirow{8}{*}{KADID-10K} & SL & 0.881 & 0.639 & 0.635 & 0.387 & 0.618 & 0.835\\
        & LwF & 0.853 & 0.736 & 0.676 & 0.418 & 0.622 & 0.842\\
        & \textcolor{black}{LwF-KG} & 0.888 & 0.731 & 0.766 & 0.741 & 0.831 & \textbf{0.849}\\
        & SI & 0.895 & 0.761 & 0.740 & 0.612 & 0.731 & 0.832\\
        & \textcolor{black}{SI-KG} & 0.871 & 0.742 & 0.782 & 0.687 & 0.764 & 0.824\\
        & MAS & 0.876 & 0.744 & 0.753 & 0.419 & 0.677 & 0.831\\
        & \textcolor{black}{MAS-KG} & 0.857 & 0.758 & 0.808 & 0.600 & 0.752 & 0.840\\
        & \textcolor{black}{TSN-IQA}  & \textcolor{black}{\textbf{0.954}} & \textcolor{black}{\textbf{0.801}} & \textcolor{black}{\textbf{0.829}} & \textcolor{black}{\textbf{0.786}} & \textcolor{black}{\textbf{0.870}} & \textcolor{black}{0.830}\\
     \bottomrule
   \end{tabular}
\end{table*}

\subsection{Experimental Setups}~\label{subsec:setup}
We select six widely used IQA datasets: LIVE~\cite{sheikh2006statistical}, CSIQ~\cite{larson2010most}, BID~\cite{ciancio2011no}, LIVE Challenge~\cite{ghadiyaram2016massive}, KonIQ-10K~\cite{hosu2020koniq}, and KADID-10K~\cite{lin2019kadid}. We summarize the details of the six datasets in Table~\ref{tab:database}. In general, the number of training pairs is proportional to the number of images in the training set of each dataset. 

Following~\cite{zhang2023continual}, we organize these datasets in chronological order for the main experiments. We randomly sample $70\%$ and $10\%$ images from each dataset for training and validation, respectively, and leave the remaining  for testing. To ensure content independence in  LIVE, CSIQ, and KADID-10K, we divide the training and test sets according to the reference images.

We choose a variant of ResNet-18 \cite{he2016deep} as the feature extractor. We keep the front convolution and four residual blocks, which are indexed by Stage $1$ to Stage $4$, respectively. We append an FC layer as the prediction head on top of the convolution response from Stage $4$, and compute the weightings  using the convolution responses from later two stages. \textcolor{black}{As such, $10,112$ learnable parameters are introduced by BN and FC layers for each new task, accounting for less than $0.18\%$ of the total network parameters. During inference, the number of centroids used in $K$-means is set to $K=128$ for each new task\footnote{Empirically, we find that TSN-IQA is insensitive to the choice of $K$.}, which introduces $98,304$ parameters for additional memory budget, accounting for about $0.88\%$ of the backbone network parameters. Putting together, the current configuration ensures that \textcolor{black}{TSN-IQA}  conforms to the bounded memory footprint desideratum.}

For each task, stochastic optimization is carried out by Adam~\cite{kingma2015adam} with an initial learning rate of $1\times 10^{-3}$. We decay the learning rate by a factor of $10$ \textcolor{black}{at the $8$-th epoch}, and train our method for a maximum of \textcolor{black}{twelve} epochs. We set the temperature to $\tau = 32$ in Eq. \eqref{eq:softmin}. We test on images of the original size. 

We use Spearman's rank correlation coefficient (SRCC) to measure the prediction performance. \textcolor{black}{When continually learning a BIQA model on a $T$-length task sequence, we compute the mean SRCC between model predictions and MOSs of each dataset as a measure of prediction accuracy:
\begin{equation}\label{eq:msrcc}
    \mathrm{mSRCC}  = \frac{1}{T}\sum_{k=1}^{T}\mathrm{SRCC}_{Tk},
\end{equation}
where $\mathrm{SRCC}_{tk}$ is the SRCC result of the $t$-th model on the $k$-th dataset. We then compute a mean plasticity index ($\mathrm{mPI}$):
\begin{equation}\label{eq:plasticity}
    \mathrm{mPI}  =\frac{1}{T}\sum_{t=1}^{T}\mathrm{PI}_{t}= \frac{1}{T}\sum_{t=1}^{T}\mathrm{SRCC}_{tt},
\end{equation}
\ie, the average result of the model on the current dataset along the task sequence, and a mean stability index ($\mathrm{mSI}$) by measuring the variability of model performance on old data when learning on a new task:
\begin{equation}\label{eq:stability1}
\mathrm{mSI} = \frac{1}{T}\sum_{t=1}^{T}\mathrm{SI}_{t},
\end{equation}
where
\begin{equation}\label{eq:stability2}
\mathrm{SI}_{t} =\begin{cases}
    1 &  t=1 \\
    \frac{1}{t-1}\sum_{k=1}^{t-1}\widehat{\mathrm{SRCC}}_{tk} & t > 1 
  \end{cases},
\end{equation}
where $\widehat{\mathrm{SRCC}}_{tk}$ for $k < t$ is computed between the predictions of the $t$-th and $k$-th models. $\mathrm{mSRCC}$, $\mathrm{mPI}$, and $\mathrm{mSI}$ measure different and complementary aspects of a continually learned BIQA model. We also quantify the plasticity-stability trade-off using a mean plasticity-stability index ($\mathrm{mPSI}$) over a list of $T$ tasks:
\begin{equation}\label{eq:psi}
\mathrm{mPSI} =\frac{1}{T}\sum_{t=1}^{T}\mathrm{PSI}_{t} =\frac{1}{2T}\sum_{t=1}^{T}\left(\mathrm{PI}_{t} + \mathrm{SI}_{t}\right).
\end{equation}}

\subsection{Competing Methods}~\label{subsec:training_methods}
We describe several competing methods for training. \textcolor{black}{For a fair comparison, we rely on the same backbone network (\ie, ResNet-18) as TSN-IQA for implementation.  We further instantiate TSN-IQA using a two-stream DNN, composed of a variant of ResNet-18 and a VGG-like network, which allows for a direct performance comparison with other methods that share the setup  in~\cite{zhang2023continual}. }

\begin{itemize}
    \item\textbf{Separate Learning (SL)} is the standard in BIQA, which trains the model using a single prediction head on one of the six training sets.
    \item\textbf{Joint Learning (JL)} refers to the dataset combination trick~\cite{zhang2021uncertainty} to address the cross-distortion-scenario challenge in BIQA. As an upper bound of all continual learning methods, JL trains the model with a single head on the combination of all six training sets. 
    
    \item \textbf{LwF}~\cite{li2018learning} in BIQA is based on a multi-head architecture, which introduces a stability regularizer that uses the previous model outputs as soft labels to preserve the performance of previously seen data. LwF relies on the newest head for quality prediction.  We also leverage the task oracle to select the corresponding head for quality prediction, denoted by \textbf{LwF-O}.
    
    \item \textcolor{black}{\textbf{LwF-KG}~\cite{zhang2023continual} uses a modified LwF~\cite{li2018learning} for training and the KG mechanism for inference.} 
    \item \textbf{SI}~\cite{zenke2017continual} is also a regularization-based continual learning method, which estimates important parameters for previous tasks.  
   
    Similar to LwF, we implement a multi-head 
     architecture for SI, and rely on the newest head to predict image quality. We try to improve the performance with the KG mechanism, denoted by \textcolor{black}{\textbf{SI-KG}}, and leverage the task oracle as well, denoted by 
     \textbf{SI-O}.

    \item \textbf{MAS}~\cite{zenke2017continual} shares a similar philosophy with SI to penalize the changes to important weights. The difference lies only in the calculation of the cumulative importance measure. 

     Similarly, MAS uses the latest head for quality prediction, and has two variants that include the \textcolor{black}{KG} module and the task oracle, denoted by \textcolor{black}{\textbf{MAS-KG}} and \textbf{MAS-O}, respectively.
 
    \item \textcolor{black}{TSN-IQA} makes use of task-specific BN to handle new tasks, and enhances the~\textcolor{black}{KG} module in \cite{zhang2023continual} using rich feature hierarchies with less memory footprint. We also replace the~\textcolor{black}{KG} module with the task oracle for quality prediction, denoted by \textbf{TSN-IQA-O}.
\end{itemize}

\begin{figure*}[t]
    \centering
    \subfloat[ResNet-18 as the backbone]{\includegraphics[width=0.42\textwidth]{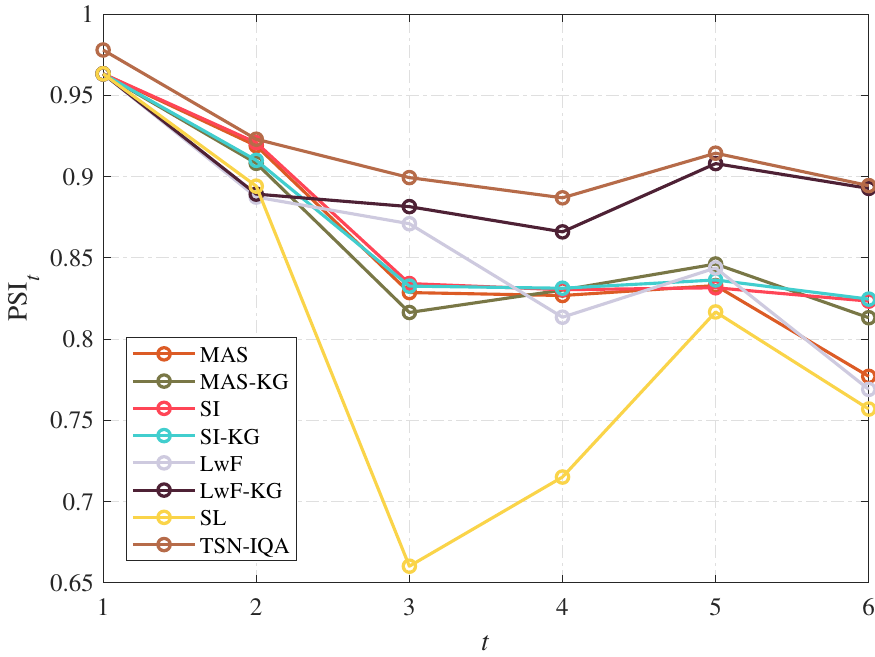}}\hskip.3em
    \subfloat[Two-stream DNN as the backbone]{\includegraphics[width=0.42\textwidth]{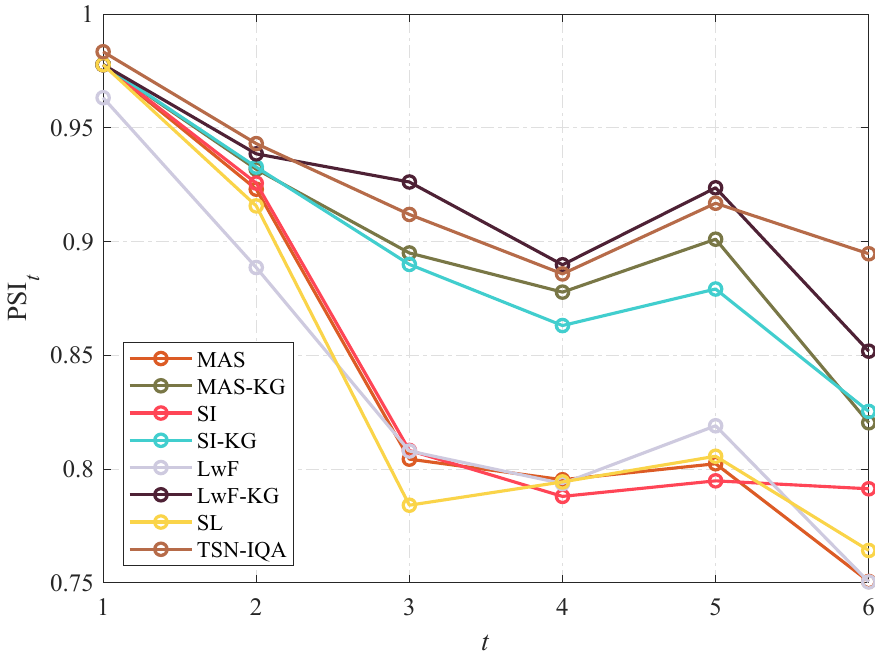}}\hskip.1em
  \caption{${\mathrm{PSI}}_t$ as a function of the task index $t$.}
\label{fig:psi}

\end{figure*}

\begin{figure*}[t]
    \centering
    \subfloat[\textcolor{black}{$\hat{q}(x) = -0.599$} ]{\includegraphics[width=0.45\textwidth]{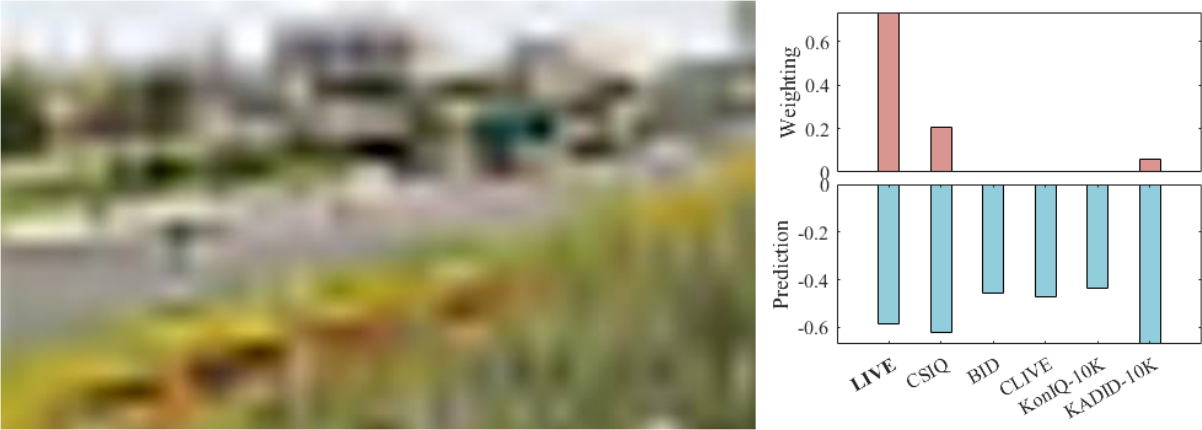}}\hskip.1em
    \subfloat[\textcolor{black}{$\hat{q}(x) = -0.480$}]{\includegraphics[width=0.45\textwidth]{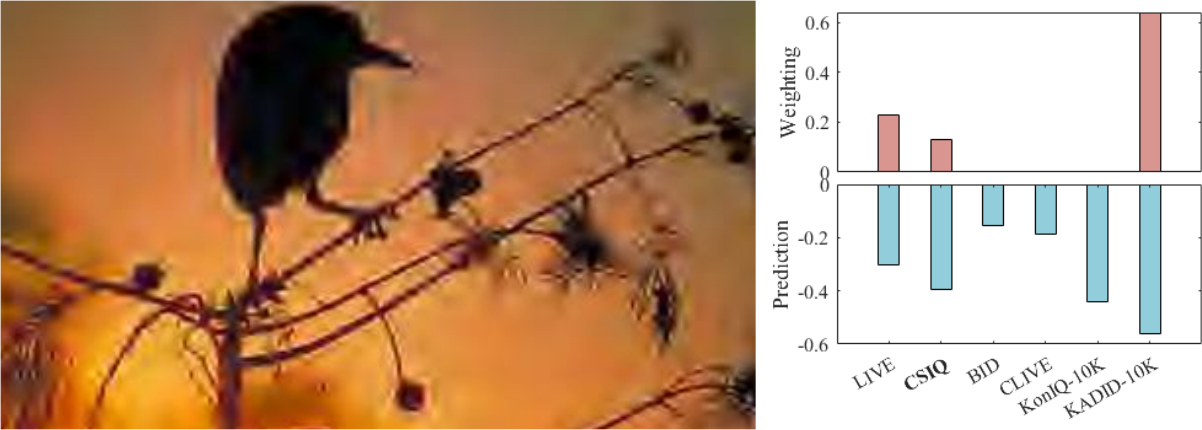}}\hskip.1em
    \subfloat[$\textcolor{black}{\hat{q}(x) = 0.353}$] {\includegraphics[width=0.45\textwidth]{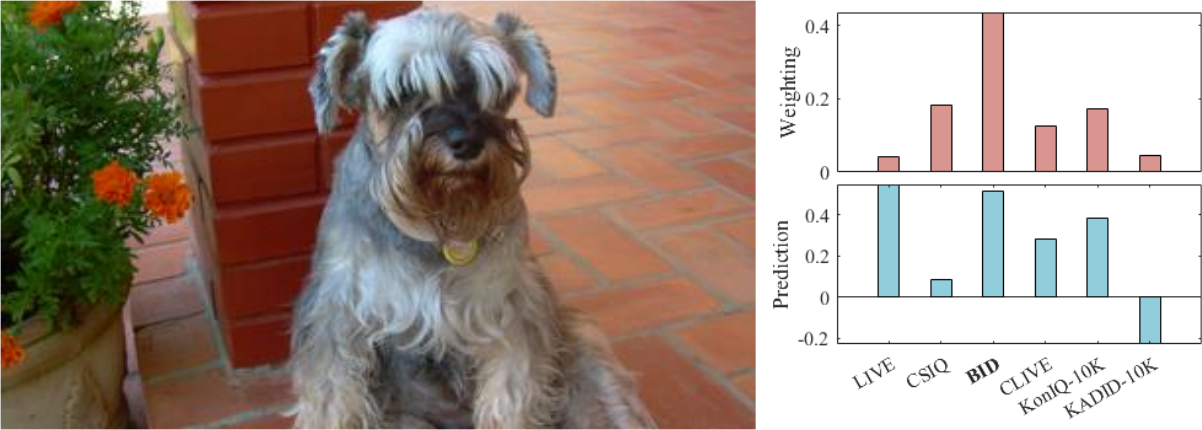}}\hskip.1em
    \subfloat[$\textcolor{black}{\hat{q}(x) = -0.394}$]{\includegraphics[width=0.45\textwidth]{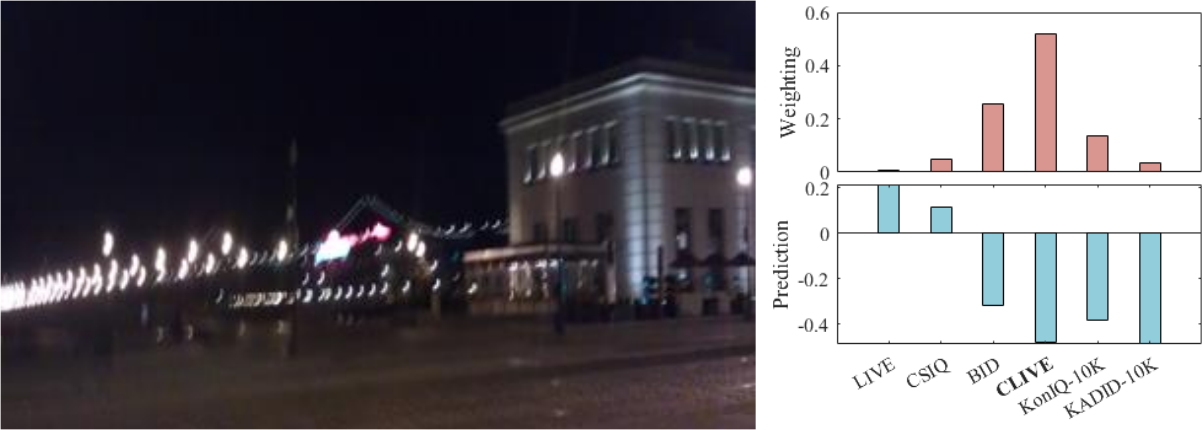}}
    \vspace{0em}
    \subfloat[$\textcolor{black}{\hat{q}(x) = -0.304}$]{\includegraphics[width=0.45\textwidth]{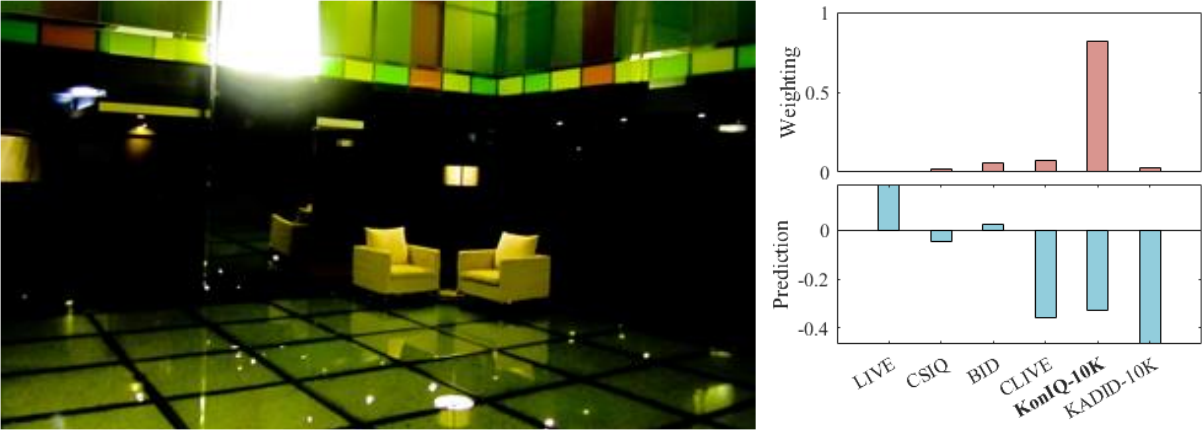}}\hskip.1em
    \subfloat[$\textcolor{black}{\hat{q}(x) = -0.618}$]{\includegraphics[width=0.45\textwidth]{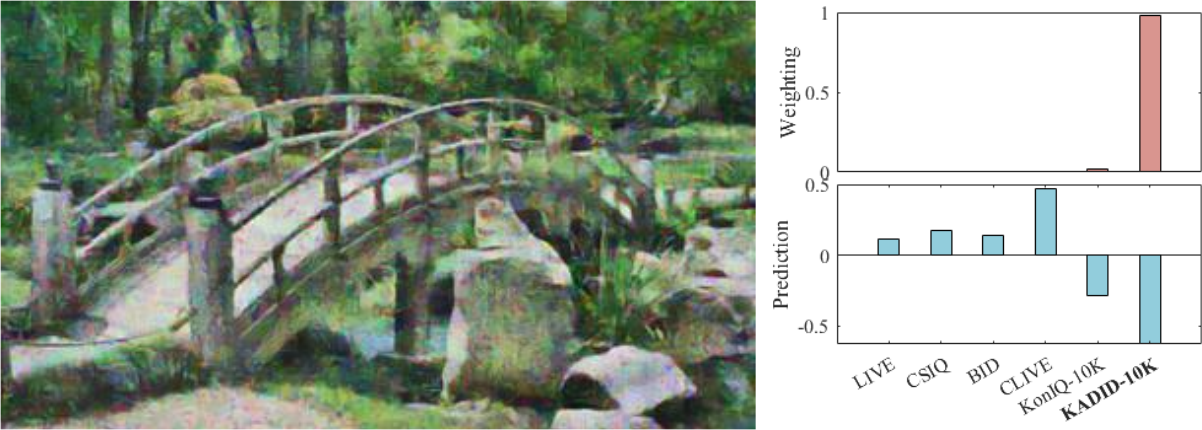}}\hskip.1em
  \caption{Learned common perceptual scale to embed images from the six IQA datasets. The bar charts of \textcolor{black}{weightings} and quality predictions are also presented alongside each image. The dataset in bold indicates the origin of the test image. The final quality prediction $\hat{q}(x)$ is shown in the subcaption. Zoom in for better distortion visibility.}
\label{fig:qualitative}
\end{figure*}

\subsection{Main Results}~\label{subsec:results}
Table~\ref{tab:mpsr} lists 
\textcolor{black}{$\mathrm{mSRCC}$, $\mathrm{mPI}$, $\mathrm{mSI}$, and $\mathrm{mPSI}$} results on the six IQA test sets. Several interesting observations have been made.
First, without any remedy for catastrophic forgetting, the performance of SL is far from satisfactory, consistent with  previous findings~\cite{zhang2023continual}. Particularly, we identify a significant performance drop when SL transits from CSIQ to BID (see Table~\ref{tab:srcc_results}), where an apparent subpopulation shift from synthetic to realistic distortions is introduced. 
Second, direct application of LwF, SI, and MAS from image classification to BIQA achieves \textcolor{black}{significantly better performance over SL in terms of $\mathrm{mSI}$ and similar performance in terms of $\mathrm{mPI}$.} 

Third, equipped with the \textcolor{black}{KG module}, LwF-KG, SI-KG, and MAS-KG outperform their counterparts \textcolor{black}{in terms of both $\mathrm{mSRCC}$ and $\mathrm{mPSI}$}. The performance is even comparable to their ``upper bounds'' (\ie, LwF-O, SI-O, and MAS-O) \textcolor{black}{under all evaluation measures. Notably, compared to \textcolor{black}{TSN-IQA}  which shares the majority of parameters between the feature extractor and the KG module, these models require twice more memory budget.}
Fourth, \textcolor{black}{TSN-IQA-O}  achieves the best results, which outperforms LwF-O, SI-O, and MAS-O in terms of \textcolor{black}{$\mathrm{mSRCC}$ and $\mathrm{mPSI}$ by large margins, and serves as the ``upper bound'' of \textcolor{black}{TSN-IQA}  for further improvement.} 
\textcolor{black}{Fifth, both LwF-KG and TSN-IQA demonstrate improved performance regarding the plasticity/stability trade-off by switching ResNet-18 to a more advanced two-stream DNN.}
\textcolor{black}{Sixth, when the task oracle is available, LwF-O is comparable to TSN-IQA-O for the two-stream backbone. But, this is not the case when ResNet-18 is the backbone, where TSN-IQA-O significantly surpasses LwF-O. This indicates that TSN-IQA is more resilient to changes in the backbone architecture.}
Lastly, TSN-IQA consistently outperforms LwF-KG~\cite{zhang2023continual} in terms of $\mathrm{mPSI}$ and $\mathrm{mSRCC}$ when using the same backbone network (ResNet-18 or the two-stream DNN), verifying the technical contribution of the proposed approach.

\begin{table*}[htbp]
  \centering
  \caption{Comparison of task-order robustness measured by $\mathrm{mSRCC}$, $\mathrm{mPI}$, $\mathrm{mSI}$, and $\mathrm{mPSI}$. \rom{1}: Default chronological order. \rom{2}: Synthetic and realistic distortions in alternation. \rom{3}: Synthetic distortions followed by realistic distortions. \rom{4}:  Realistic distortions followed by synthetic distortions. \rom{5} to \rom{8}: Reverses of Orders \rom{1} to \rom{4}, respectively}\label{tab:robustness_o}
  \begin{tabular}{l|cc|cc|cc|cc}
      \toprule
      {Order} & \multicolumn{2}{c|}{\rom{1}}  & \multicolumn{2}{c|}{\rom{2}}  & \multicolumn{2}{c|}{\rom{3}}  & \multicolumn{2}{c}{\rom{4}}\\
     \hline
     {Metric} &LwF-KG & \textcolor{black}{TSN-IQA}  &LwF-KG & \textcolor{black}{TSN-IQA}   &LwF-KG & \textcolor{black}{TSN-IQA}   &LwF-KG & \textcolor{black}{TSN-IQA} \\
     \hline
    $\mathrm{mSRCC}$ & 0.801 & \textbf{0.846} & 0.807 & \textbf{0.846}  & 0.832 & \textbf{0.846}  & 0.781 & \textbf{0.846}\\
    $\mathrm{mPI}$ & 0.837 & \textbf{0.853} & 0.836 & \textbf{0.855}  & 0.836 & \textbf{0.852}  & 0.819 & \textbf{0.850}\\
    $\mathrm{mSI}$ & 0.963 & \textbf{0.979} & 0.957 & \textbf{0.986}  & 0.958 & \textbf{0.979}  & 0.955 & \textbf{0.980}\\
    $\mathrm{mPSI}$ & 0.900 & \textbf{0.916} & 0.897 & \textbf{0.921}  & 0.897 & \textbf{0.916}  & 0.887 & \textbf{0.915}\\
    \midrule
      {Order} & \multicolumn{2}{c|}{\rom{5}}  & \multicolumn{2}{c|}{\rom{6}}  & \multicolumn{2}{c|}{\rom{7}}  & \multicolumn{2}{c}{\rom{8}}\\
     \hline
     {Metric} &LwF-KG & \textcolor{black}{TSN-IQA}  &LwF-KG & \textcolor{black}{TSN-IQA}   &LwF-KG & \textcolor{black}{TSN-IQA}   &LwF-KG & \textcolor{black}{TSN-IQA} \\
     \hline
    $\mathrm{mSRCC}$ & 0.803 & \textbf{0.846} & 0.803 & \textbf{0.846}  & 0.771 & \textbf{0.846}  & 0.808 & \textbf{0.846}\\
    $\mathrm{mPI}$ & 0.841 & \textbf{0.847} & 0.834 & \textbf{0.850}  & 0.843 & \textbf{0.849}  & 0.829 & \textbf{0.845}\\
    $\mathrm{mSI}$ & 0.959 & \textbf{0.979} & 0.958 & \textbf{0.989}  & 0.937 & \textbf{0.990}  & 0.962 & \textbf{0.976}\\
    $\mathrm{mPSI}$ & 0.900 & \textbf{0.913} & 0.896 & \textbf{0.920}  & 0.890 & \textbf{0.920}  & 0.896 & \textbf{0.911}\\
     \bottomrule
  \end{tabular}
\end{table*}

\begin{table*}[htbp]
  \centering
  \caption{Comparison of task-length robustness measured by  $\mathrm{mPSI}$ for different task orders }\label{tab:robustness_l}
  \begin{tabular}{c|cc|cc|cc|cc}
      \toprule
      {Order} & \multicolumn{2}{c|}{\rom{1}}  & \multicolumn{2}{c|}{\rom{2}}  & \multicolumn{2}{c|}{\rom{3}}  & \multicolumn{2}{c}{\rom{4}}\\
     \hline
     {Length} &LwF-KG & \textcolor{black}{TSN-IQA}  &LwF-KG & \textcolor{black}{TSN-IQA}   &LwF-KG & \textcolor{black}{TSN-IQA}   &LwF-KG & \textcolor{black}{TSN-IQA} \\
     \hline
    1 & 0.963 & 0.978 & 0.963 & 0.978  & 0.963 & 0.978  & 0.882 & 0.906\\
    2 & 0.926 & 0.950 & 0.929 & 0.944  & 0.926 & 0.950  & 0.886 & 0.900\\
    3 & 0.911 & 0.933 & 0.911 & 0.937  & 0.915 & 0.934  & 0.895 & 0.907\\
    4 & 0.900 & 0.922 & 0.900 & 0.926  & 0.903 & 0.925  & 0.901 & 0.923\\
    5 & 0.902 & 0.920 & 0.898 & 0.921  & 0.895 & 0.916  & 0.889 & 0.917\\
    6 & 0.900 & 0.916 & 0.897 & 0.921  & 0.898 & 0.916  & 0.887 & 0.915\\
    \hline
    Mean & 0.917 & \textbf{0.937} & 0.916 & \textbf{0.938} & 0.897 & \textbf{0.936} & 0.890 & \textbf{0.911}\\
    \midrule
      {Order} & \multicolumn{2}{c|}{\rom{5}}  & \multicolumn{2}{c|}{\rom{6}}  & \multicolumn{2}{c|}{\rom{7}}  & \multicolumn{2}{c}{\rom{8}}\\
     \hline
     {Length} &LwF-KG & \textcolor{black}{TSN-IQA}  &LwF-KG & \textcolor{black}{TSN-IQA}   &LwF-KG & \textcolor{black}{TSN-IQA}   &LwF-KG & \textcolor{black}{TSN-IQA} \\
     \hline
    1 & 0.918 & 0.912 & 0.946 & 0.941  & 0.946 & 0.941  & 0.918 & 0.912\\
    2 & 0.920 & 0.918 & 0.918 & 0.924  & 0.923 & 0.919  & 0.884 & 0.906\\
    3 & 0.903 & 0.906 & 0.904 & 0.913  & 0.912 & 0.914  & 0.896 & 0.923\\
    4 & 0.903 & 0.904 & 0.886 & 0.911  & 0.903 & 0.912  & 0.902 & 0.923\\
    5 & 0.893 & 0.902 & 0.887 & 0.909  & 0.887 & 0.910  & 0.894 & 0.913\\
    6 & 0.900 & 0.913 & 0.896 & 0.919  & 0.890 & 0.920  & 0.896 & 0.911\\
    \hline
    Mean & 0.906 & \textbf{0.909} & 0.906 & \textbf{0.920} & 0.910 & \textbf{0.919} & 0.898 & \textbf{0.915}\\
     \bottomrule
  \end{tabular}
\end{table*}

\begin{table}[htbp]
  \centering
  \caption{Performance comparison of \textcolor{black}{TSN-IQA}  with different design choices. \textcolor{black}{The results of LwF-KG with different backbone networks are listed as reference.} The default setting is highlighted in bold}\label{tab:mpsr_variant}
  \begin{tabular}{l|l|cc}
      \toprule
    \multicolumn{2}{l|}{Design Choice} &  $\mathrm{mPSI}$ & $\mathrm{mSRCC}$ \\
     \hline
    \multicolumn{2}{l|}{LwF-KG (ResNet-18)} & 0.900 & 0.801 \\
     \hline
    \multicolumn{2}{l|}{LwF-KG (Two-Stream DNN)} & 0.918 & 0.815 \\
    \hline
    \multicolumn{2}{l|}{Task-agnostic BN} & 0.822 & 0.624\\
    \hline
    \multicolumn{2}{l|}{ImageNet Pre-trained BN} & 0.893 & 0.835\\
    \hline
    & Stage 4  & 0.914 & 0.839\\
    Feature &Stages \textbf{3+4}   & \textbf{0.916} & \textbf{0.846}\\
    Hierarchy &Stages 2+3+4  & 0.914 & 0.845\\
    & Stages 1+2+3+4  & 0.913 & 0.844\\

    \hline
     & VGG-16  & 0.920 & 0.845\\
    Backbone & ResNet-50  & \textbf{0.924} & 0.845\\
     & Two-Stream DNN & 0.923 & \textbf{0.846}\\
     \bottomrule
  \end{tabular}
\end{table}

\begin{table}[htbp]
  \centering
  \caption{IQA dataset analysis via pairwise Kullback–Leibler (KL) divergence of task-specific BN parameters. A lower value indicates the two datasets are more similar}\label{tab:kl_div}
  \setlength{\tabcolsep}{1.5mm}\begin{tabular}{l|cccccc}
      \toprule
    Dataset & LIVE & CSIQ & BID & CLIVE & KonIQ & KADID\\
    \hline
    LIVE & 0.000 & 84.680 & 131.015 & 144.343 & 188.037
    & 135.388 \\
    CSIQ & 81.546 & 0.000 & 142.367 & 152.962 & 218.956 & 144.279  \\
    BID & 101.814 & 111.335 & 0.000 & 63.436 & 132.474 & 153.364 \\
    CLIVE &123.188 & 124.207 & 64.462 & 0.000 & 162.499 & 170.714  \\
    KonIQ & 110.939 & 126.010 & 111.538 & 122.768 & 0.000 & 138.842  \\
    KADID & 93.250 & 102.867 & 134.079 & 141.610 & 151.447 & 0.000 \\
     \bottomrule
   \end{tabular}
\end{table}

\begin{table}[htbp]
  \centering
  \caption{SRCC results of intra- and inter-dataset evaluations. ``Inter-'' means the set of BN parameters (together with the corresponding prediction head) trained for one task are tested on other tasks}\label{tab:srcc_div}
  \begin{tabular}{l|cccccc}
      \toprule
    Dataset & LIVE & CSIQ & BID & CLIVE & KonIQ & KADID\\
    \hline
    LIVE & 0.956 & 0.648 & 0.680 & 0.453 & 0.665 & 0.529 \\
    CSIQ & 0.893 & 0.852 & 0.560 & 0.383 & 0.512 & 0.530 \\
    BID & 0.682 & 0.751 & 0.812 & 0.691 & 0.698 & 0.496 \\
    CLIVE & 0.534 & 0.489 & 0.803 & 0.807 & 0.747 & 0.441  \\
    KonIQ & 0.687 & 0.563 & 0.721 & 0.646 & 0.881 & 0.601  \\
    KADID & 0.916 & 0.693 & 0.618 & 0.436 & 0.593 & 0.824 \\
     \bottomrule
   \end{tabular}
\end{table}

\begin{table*}[htbp]
  \centering
  \caption{SRCC results on  TID2013~\cite{ponomarenko2013color}, SPAQ~\cite{fang2020perceptual}, \textcolor{black}{and AGIQA-3K~\cite{li2023agiqa}} datasets along with the mean weightings for all prediction heads. \textcolor{black}{A variant of ResNet-18 is used as the backbone network}}\label{tab:cross}
  \begin{tabular}{l|cccccc|c}
      \toprule
    \multirow{2}{*}{Dataset} & \multicolumn{6}{c|}{Mean Weightings} &\multirow{2}{*}{SRCC}\\
    \cline{2-7}
    & LIVE & CSIQ & BID & LIVE Challenge & KonIQ-10K & KADID-10K & \\
    \hline
    TID2013 & 0.241 & 0.138 & 0.042 & 0.147 & 0.114 & 0.318 & 0.700\\
    SPAQ  & 0.076 & 0.092 & 0.146 & 0.206 & 0.377 & 0.102 & 0.817\\
    \textcolor{black}{AGIQA-3K}  & \textcolor{black}{0.141} & \textcolor{black}{0.106} & \textcolor{black}{0.050} & \textcolor{black}{0.437} & \textcolor{black}{0.135} & \textcolor{black}{0.130} & \textcolor{black}{0.615}\\
     \bottomrule
   \end{tabular}
\end{table*}

\begin{table}[tp]
  \centering
  \caption{\textcolor{black}{SRCC results on  TID2013~\cite{ponomarenko2013color}, SPAQ~\cite{fang2020perceptual}, and AGIQA-3K~\cite{li2023agiqa} datasets under the cross-dataset setup. All methods use a variant of ResNet-50 as the backbone network}}\label{tab:cross2}
  \begin{tabular}{l|ccc}
      \toprule
    \textcolor{black}{Method} & \textcolor{black}{TID2013} & \textcolor{black}{SPAQ} & \textcolor{black}{AGIQA-3K}\\
    \hline
     \textcolor{black}{PQR~\cite{zeng2018blind}} & \textcolor{black}{0.528} & \textcolor{black}{0.808} & \textcolor{black}{0.642}\\
     \textcolor{black}{HyperIQA~\cite{9156687}} & \textcolor{black}{0.482} & \textcolor{black}{0.772} & \textcolor{black}{0.628} \\
    \textcolor{black}{CONTRIQUE~\cite{madhusudana2022image}} & \textcolor{black}{0.318} & \textcolor{black}{0.637} & \textcolor{black}{0.601} \\
    \hline
    \textcolor{black}{TSN-IQA (ResNet-50)} & \textbf{0.734} & \textbf{0.821} & \textbf{0.648}\\
     \bottomrule
   \end{tabular}
\end{table}

\begin{table}[htbp]
  \centering
  \caption{Performance comparison of TSN-IQA with different KG mechanisms in terms of $\mathrm{mSRCC}$, $\mathrm{mPSI}$, and the computational complexity (CC). $T$ stands for the number of learned tasks}\label{tab:balance}
  \begin{tabular}{l|ccc}
      \toprule
    {Configuration} & $\mathrm{mSRCC}$ & $\mathrm{mPSI}$ & CC\\
    \hline 
     LwF-KG (as baseline) & 0.801 & 0.900 & 1.0\\
     \hline
     TSN-IQA (with soft assignment) & 0.846 & 0.916 & $T$\\
     TSN-IQA (with hard assignment) & 0.817 & 0.903 & 1.0\\ 
     \bottomrule
  \end{tabular}
\end{table}

We plot ${\mathrm{PSI}}_{t}$ as a function of the task index $t$ in Fig.~\ref{fig:psi}, from which we \textcolor{black}{find our method is more stable as the length of the task sequence grows for different backbone networks.}  We then look closely at the performance variations along the task sequence, and summarize the SRCC results continually in Table~\ref{tab:srcc_results}. 

Several useful findings are worth mentioning.
First, JL provides an effective but unscalable solution to the subpopulation shift in BIQA, serving as the upper bound of all continual learning methods. 
Second, the plasticity of SL is \textcolor{black}{reasonably good}, but the results \textcolor{black}{on previously learned tasks} suffer from significant oscillations due to the subpopulation shift between synthetic and realistic distortions~\cite{zeng2018blind,zhang2020blind,zhang2021uncertainty}. 
Third, the favorable performance of \textcolor{black}{LwF-KG}, \textcolor{black}{SI-KG}, and \textcolor{black}{MAS-KG} against LwF, SI, and MAS especially on old tasks validates the \textcolor{black}{KG module} for summarizing quality predictions.

\begin{table*}[t]
  \centering
  \caption{Mean results (and the corresponding standard deviations) of TSN-IQA against other methods over five random initializations }\label{tab:random_seed}
  \begin{tabular}{c|l|cccc}
      \toprule
     Backbone & {Method} & $\mathrm{mSRCC}$ & $\mathrm{mPI}$ & $\mathrm{mSI}$ & $\mathrm{mPSI}$\\
    \hline
     \multirow{8}{*}{ResNet-18}& SL & {0.695} ({\color{grey}$\pm$ 0.018}) & {0.852} ({\color{grey}$\pm$ 0.005}) & {0.753} ({\color{grey}$\pm$ 0.004}) & {0.802} ({\color{grey}$\pm$ 0.003}) \\
    & SI & {0.677} ({\color{grey}$\pm$ 0.116})  & \textbf{0.856} ({\color{grey}$\pm$ 0.007})  & {0.868} ({\color{grey}$\pm$ 0.010})  & {0.862} ({\color{grey}$\pm$ 0.008})  \\
    & SI-KG & {0.712} ({\color{grey}$\pm$ 0.086})  & {0.854} ({\color{grey}$\pm$ 0.008})  & {0.866} ({\color{grey}$\pm$ 0.019})  & {0.860} ({\color{grey}$\pm$ 0.013}) \\
   &  MAS & {0.717} ({\color{grey}$\pm$ 0.018}) & {0.854} ({\color{grey}$\pm$ 0.007}) & {0.865} ({\color{grey}$\pm$ 0.006}) & {0.859} ({\color{grey}$\pm$ 0.006}) \\
    & MAS-KG & {0.747} ({\color{grey}$\pm$ 0.022}) & {0.853} ({\color{grey}$\pm$ 0.006}) & {0.871} ({\color{grey}$\pm$ 0.007}) & {0.862} ({\color{grey}$\pm$ 0.006})\\
    & LwF & {0.697}  ({\color{grey}$\pm$ 0.026}) & {0.846}  ({\color{grey}$\pm$ 0.004}) & {0.880}  ({\color{grey}$\pm$ 0.011}) & {0.863}  ({\color{grey}$\pm$ 0.007}) \\
    & LwF-KG & {0.801} ({\color{grey}$\pm$ 0.003}) & {0.840} ({\color{grey}$\pm$ 0.003}) & {0.964} ({\color{grey}$\pm$ 0.003}) & {0.902} ({\color{grey}$\pm$ 0.002}) \\
     &{TSN-IQA}  & \textbf{0.839} ({\color{grey}$\pm$ 0.004}) & {0.851} ({\color{grey}$\pm$ 0.003}) & \textbf{0.983} ({\color{grey}$\pm$ 0.003}) & \textbf{0.917} ({\color{grey}$\pm$ 0.001})\\
     \midrule
     Backbone & {Method} & $\mathrm{mSRCC}$ & $\mathrm{mPI}$ & $\mathrm{mSI}$ & $\mathrm{mPSI}$\\
    \hline
     \multirow{8}{*}{Two-Stream DNN}& SL & {0.672} ({\color{grey}$\pm$ 0.016}) & {0.875} ({\color{grey}$\pm$ 0.005}) & {0.802} ({\color{grey}$\pm$ 0.008}) & {0.839} ({\color{grey}$\pm$ 0.004}) \\
    & SI & {0.705}  ({\color{grey}$\pm$ 0.021}) & {0.875}  ({\color{grey}$\pm$ 0.005}) & {0.833}  ({\color{grey}$\pm$ 0.008}) & {0.854}  ({\color{grey}$\pm$ 0.005})\\
    & SI-KG & {0.799}  ({\color{grey}$\pm$ 0.008}) & {0.862}  ({\color{grey}$\pm$ 0.009}) & {0.948}  ({\color{grey}$\pm$ 0.011}) & {0.905}  ({\color{grey}$\pm$ 0.005})\\
   &  MAS & {0.687} ({\color{grey}$\pm$ 0.043}) & \textbf{0.877} ({\color{grey}$\pm$ 0.003}) & {0.826} ({\color{grey}$\pm$ 0.020}) & {0.851} ({\color{grey}$\pm$ 0.011}) \\
    & MAS-KG & {0.806} ({\color{grey}$\pm$ 0.015}) & {0.865} ({\color{grey}$\pm$ 0.003}) & {0.950} ({\color{grey}$\pm$ 0.006}) & {0.908} ({\color{grey}$\pm$ 0.004})\\
    & LwF & {0.688}  ({\color{grey}$\pm$ 0.019}) & {0.873}  ({\color{grey}$\pm$ 0.011}) & {0.841}  ({\color{grey}$\pm$ 0.008}) & {0.857}  ({\color{grey}$\pm$ 0.009}) \\
    & LwF-KG & {0.810} ({\color{grey}$\pm$ 0.009}) & {0.850} ({\color{grey}$\pm$ 0.008}) & {0.979} ({\color{grey}$\pm$ 0.001}) & {0.914} ({\color{grey}$\pm$ 0.004}) \\
     &{TSN-IQA}  & \textbf{0.839} ({\color{grey}$\pm$ 0.009}) & {0.858} ({\color{grey}$\pm$ 0.005}) & \textbf{0.981} ({\color{grey}$\pm$ 0.004}) & \textbf{0.920} ({\color{grey}$\pm$ 0.004})\\
     \bottomrule
  \end{tabular}
\end{table*}

We last conduct a qualitative analysis of our BIQA model by showing representative test images from the task sequence in Fig.~\ref{fig:qualitative}. We find that for frequently-seen distortion appearances (\eg, global blurring in  (a)), all heads tend to make reasonable predictions, and more weightings are given to the corresponding head. Meanwhile, if one distortion type occurs in multiple datasets (\eg, JPEG2000 compression in (b)), the heads that have seen the distortion work well, while others do not. Fortunately, the \textcolor{black}{KG module} is able to underweight inaccurate heads. Moreover, TSN-IQA successfully aligns images of synthetic and realistic distortions in a common perceptual scale, despite not being exposed to pairs of images from different distortion scenarios.

\subsection{Results of Task-Order/-Length Robustness}~\label{subsec:robustness}
In real-world applications, novel distortions may emerge in arbitrary order. As a result, a continual learning method for BIQA is expected to be robust to different task orders. In addition to (\rom{1}) the default chronological order, we experiment with seven more task orders: (\rom{2}) synthetic and realistic distortions in alternation: LIVE $\rightarrow$ BID $\rightarrow$ CSIQ $\rightarrow$ LIVE Challenge $\rightarrow$ KADID-10K $\rightarrow$ KonIQ-10K, (\rom{3}) synthetic distortions followed by realistic distortions: LIVE $\rightarrow$ CSIQ $\rightarrow$ KADID-10K $\rightarrow$ BID $\rightarrow$ LIVE Challenge $\rightarrow$ KonIQ-10K, (\rom{4}) realistic distortions followed by synthetic distortions: BID $\rightarrow$ LIVE Challenge $\rightarrow$ KonIQ-10K $\rightarrow$ LIVE $\rightarrow$ CSIQ $\rightarrow$ KADID-10K, and (\rom{5})-(\rom{8}) the reverses of Orders (\rom{1})-(\rom{4}). 

We compare \textcolor{black}{TSN-IQA} to \textcolor{black}{LwF-KG} \cite{zhang2023continual} in Table \ref{tab:robustness_o}. The main observation is that our method is more robust than \textcolor{black}{LwF-KG for all task orders under all metrics}. \textcolor{black}{Furthermore, we note that the results of Orders~\rom{5} and \rom{8} are slightly lower than those of other orders}. We believe these arise because we begin with KADID-10K \cite{lin2019kadid}, a synthetic dataset that is considered visually much harder than LIVE \cite{sheikh2006statistical} and CSIQ \cite{larson2010most}, therefore posing a challenge for performance stabilization. Given a specific task order, we also measure the task-length robustness by the mean \textcolor{black}{$\mathrm{mPSI}$} of different lengths, \textcolor{black}{$\{\mathrm{mPSI}_{t}\}_{t=1}^T$}. We compare our method to LwF-KG \cite{zhang2023continual} in Table \ref{tab:robustness_l}. We find the task-length robustness to be dependent on the task order, and \textcolor{black}{TSN-IQA} performs better than \textcolor{black}{LwF-KG} across all task orders. A relatively inferior result is observed for Order~\rom{5}, where KADID-10K \cite{lin2019kadid} is listed in the first place. Altogether, these promising results indicate that TSN-IQA has great potential for use in practical quality prediction scenarios.

\subsection{Ablation Studies}\label{subsec:ablation}
In this subsection, we conduct  ablation experiments to probe the performance variations of TSN-IQA. Note that all experiments are conducted using the default chronological order. First, to verify the necessity of the core design of our method - task-specific BN, we train a single group of task-agnostic BN parameters along the task sequence. During inference, we use the converged BN parameters to make predictions for all tasks. As shown in Table \ref{tab:mpsr_variant}, this variant achieves an \textcolor{black}{$\mathrm{mPSI}$ of $0.822$ and an $\mathrm{mSRCC}$ of $0.624$}, which are far below the results by \textcolor{black}{TSN-IQA}. \textcolor{black}{We next compare the performance using the ImageNet pre-trained BN  with the proposed distortion-aware BN  for the KG module. Being exposed to various types of distortions, the distortion-aware BN parameters help the KG module assign weightings to predictions heads more reasonably, leading to higher $\mathrm{mPSI}$ and  $\mathrm{mSRCC}$ results.} We then evaluate the influence of the feature hierarchy on the \textcolor{black}{KG module} by incorporating different stages of convolution responses. The results in Table \ref{tab:mpsr_variant} show that multi-stage features are more beneficial, and the combination of Stage-3 and Stage-4 features delivers the most perceptual gains. 

\textcolor{black}{Lastly, we experiment with three different DNNs as the backbone networks, \ie, VGG-16~\cite{simonyan2014very}, ResNet-50~\cite{he2016deep}, and the two-stream DNN~\cite{zhang2023continual}. From Table~\ref{tab:mpsr_variant}, we observe that the proposed parameter decomposition scheme is generic for continual learning of BIQA models, which can be enhanced by working with  more powerful backbone networks.}

\textcolor{black}{\subsection{Further Analysis}\label{subsec:analysis}}
\subsubsection{IQA Dataset Analysis}\label{subsubsec:correlation_bn}
\textcolor{black}{During continual learning on a task sequence, TSN-IQA is trained to capture the informative and discriminative information of each task using a group of task-specific BN parameters. It remains to be seen 1) whether the learned BN parameters reflect distinctive aspects of different datasets, and 2) whether they can explain the performance variations. To answer these questions, we first retrieve the exponentially decaying moving averages of the mean and std parameters from the last BN layer learned for each task, which are assumed to follow a multivariate Gaussian distribution. With such $T$ Gaussian distributions at hand, we compute the pairwise Kullback–Leibler (KL) divergence $\{\mathrm{KL}_{ij}\}_{i,j=1}^{T}$. From Table~\ref{tab:kl_div},  we identify a clear trend that datasets with similar distortion scenarios have  relatively smaller divergence values.  We then load each group of task-specific BN parameters (together with the corresponding prediction head), and test it on all datasets, by which we obtain pairwise SRCC results among all datasets $\{\mathrm{SRCC}\}_{i,j=1}^{T}$ (see Table~\ref{tab:srcc_div}). Finally, we are able to measure the correlation between the learned BN parameters and the performance variations with an SRCC of $-0.776$ between $\{\mathrm{SRCC}\}_{i,j=1}^{T}$ and  $\{\mathrm{KL}\}_{i,j=1}^{T}$. This provides empirical evidence that the more similar the datasets are in distortion scenarios, the better the inter-dataset prediction accuracy.
}

\subsubsection{Generalizability Analysis}\label{subsubsec:unseen}
To empirically verify that TSN-IQA can be used to predict the perceptual quality of images beyond all seen datasets, we test the model learned in chronological order of the six tasks on \textcolor{black}{three} additional datasets, including TID2013~\cite{ponomarenko2013color}, SPAQ~\cite{fang2020perceptual} and \textcolor{black}{AGIQA-3K~\cite{li2023agiqa}}. We report the SRCC results and the average weightings computed by the KG module over all images in Table~\ref{tab:cross}, from which we have two useful observations. First, the proposed TSN-IQA presents reasonable generalizability to the tasks it is not exposed to. Second, the KG module assigns perceptually meaningful weightings to the prediction heads. Specifically, the prediction heads learned on LIVE, CSIQ, and KADID-10K are assigned  higher weightings when handling TID2013, containing multiple synthetic distortions. Similarly, the prediction heads for BID, LIVE Challenge, and KonIQ-10K are assigned higher weightings for SPAQ, which is dominated by realistic camera distortions. \textcolor{black}{The situation becomes a little intricate on AGIQA-3K, a new dataset comprising artificially generated images. All prediction heads are assigned  non-trivial weightings, exposing the uncertainty of TSN-IQA in handling this novel image type. We also compare TSN-IQA with three recent BIQA methods, \ie, PQR~\cite{zeng2018blind}, HyperIQA~\cite{9156687}, and
CONTRIQUE~\cite{madhusudana2022image}, following the same cross-dataset evaluation setup. 
As shown in Table~\ref{tab:cross2}, TSN-IQA outperforms other BIQA models on all three datasets, especially on TID2013~\cite{ponomarenko2013color} that contains synthetic distortions. 
Nevertheless, there remains considerable room for improvement in quality prediction of artificially-generated images that exhibit substantial subpopulation shift.}

\subsubsection{Computational Complexity Analysis}\label{subsubsec:computational complexity}
\textcolor{black}{We compare the computational complexity of TSN-IQA with LwF-KG~\cite{zhang2023continual}. The computation of a single forward pass for the two methods are identical. Specifically, given an image with a size of $224 \times 224 \times 3$, the number of multiply–accumulate operations (MACs) of TSN-IQA is about $18.2$ G. After continually learned on $T$ tasks, TSN-IQA computes $T$ quality estimates with $T$ groups of task-specific BN parameters during inference. As such, the computational complexity is linear with respect to the number of training tasks, which can be straightforwardly accelerated by parallel computing.} 

\textcolor{black}{We have also tried a variant of the KG module that implements hard assignment by setting $\tau = +\infty$. With such a modification, only one forward pass is needed to compute the final quality score, which reduces the computational complexity by a factor of $T$. As shown in Table~\ref{tab:balance}, although this computationally efficient variant delivers slightly inferior performance than the default TSN-IQA, it outperforms LwF-KG with the same computational complexity.}

\subsubsection{\textcolor{black}{Performance Stability}}\label{subsubsec:random_seeds}
\textcolor{black}{We test the performance stability of TSN-IQA with respect to five different random initializations. As shown in Table~\ref{tab:random_seed}, TSN-IQA consistently demonstrates superior performance over other methods irrespective of the chosen backbone network.}

\section{Conclusion and Discussion}
\label{sec:conclusion}
We have introduced a simple yet effective method of continually learning BIQA models. The key to the success of the proposed TSN-IQA is to train task-specific BN parameters for each task while holding all pre-trained convolution filters fixed. \textcolor{black}{On the one hand, TSN-IQA encourages more effective feature representation learning for different tasks. This is because BN participates all network stages of feature processing, which is better suited in the continual learning scenario for noticeably improved quality prediction accuracy, plasticity-stability trade-off, and task-length/-order robustness. On the other hand, it permits a significant reduction in the number of parameters used for the KG mechanism, which only needs to replace a set of BN parameters.}

 TSN-IQA relies on five desiderata as specified in \cite{zhang2023continual}, among which the assumption of a common perceptual scale is foremost. It is well-known that the perceived quality of a visual image depends not only on the image content itself, but also on the subjective testing protocols as well as viewing conditions. For example, switching from single-stimulus methods to 2AFC approaches generally improves the accuracy of fine-grained quality annotations. We take this into consideration by pursuing binary labels as ground-truths. Moreover, the visibility of some distortions (\eg, JPEG compression) varies with the effective viewing distance. Although it would be ideal to give a complete treatment of viewing conditions (\eg, as part of the model input), our computational study shows the possibility to learn a common perceptual scale for different IQA datasets with MOSs collected under similar viewing conditions and having overlapping quality ranges. With the explosive growth of user-generated and \textcolor{black}{computer-generated} images, it is also desirable to perform online continual learning for BIQA, where there is no distinct boundaries between tasks (or datasets) during training. 

\ifCLASSOPTIONcompsoc
  \section*{Acknowledgments}
\else
  \section*{Acknowledgment}
\fi
This work was supported in part by Shanghai Municipal Science and Technology Major Project (2021SHZDZX0102), the Fundamental Research Funds for the Central Universities, the National Natural Science Foundation of China under Grants 61901262, 62371283, 62071407,  and U19B2035, and the Hong Kong RGC Early Career Scheme (9048212).

\bibliographystyle{IEEEtran}
\bibliography{Weixia}

\begin{IEEEbiography}[{\includegraphics[width=1in,height=1.25in,clip,keepaspectratio]{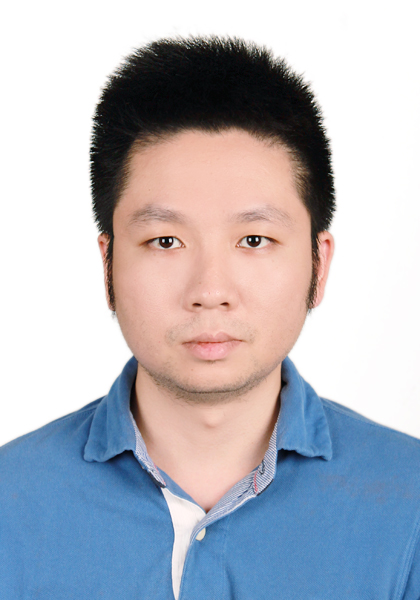}}]{Weixia Zhang}(Member, IEEE)
received the B.E. degree from the Wuhan University, Wuhan, China, in 2011 and the M.S. degree in electrical and computer engineering from the University of Rochester, NY, USA, in 2013. He then received the Ph.D. degree from the Wuhan University, Wuhan, China, in 2018. He is currently an Associate Research Scientist with the MoE Key Lab of Artificial Intelligence, AI Institute, Shanghai Jiao Tong University. His research interests include perceptual quality evaluation and enhancement.
\end{IEEEbiography}

\begin{IEEEbiography}[{\includegraphics[width=1in,height=1.25in,clip,keepaspectratio]{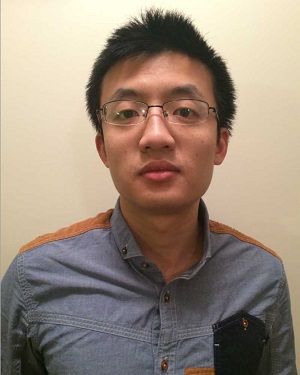}}]{Kede Ma} (Senior Member, IEEE) received the B.E. degree from the University of Science and Technology of China, Hefei, China, in 2012, and the M.S. and Ph.D. degrees in electrical and computer engineering from the University of Waterloo, Waterloo, ON, Canada, in 2014 and 2017, respectively. He was a Research Associate with the Howard Hughes Medical Institute and New York University, New York, NY, USA, in 2018. He is currently an Assistant Professor with the Department of Computer Science, City University of Hong Kong. His research interests include perceptual image processing, computational vision, and computational photography.
\end{IEEEbiography}

\begin{IEEEbiography}[{\includegraphics[width=1in,height=1.25in,clip,keepaspectratio]{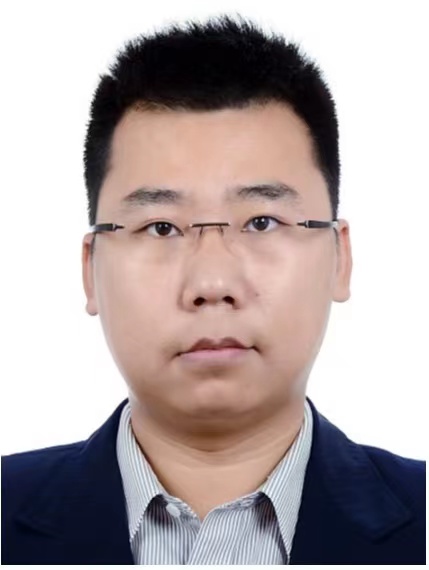}}]{Guangtao Zhai}(Senior Member, IEEE) received the B.E. and M.E. degrees from Shandong University, Shandong, China, in 2001 and 2004, respectively, and the Ph.D. degree from Shanghai Jiao Tong University, Shanghai, China, in 2009, where he is currently a Research Professor with the Institute of Image Communication and Information Processing. From
2008 to 2009, he was a Visiting Student with the Department of Electrical and Computer Engineering, McMaster University, Hamilton, ON, Canada, where he was a Post-Doctoral Fellow from 2010 to 2012. From 2012 to 2013, he was a Humboldt Research Fellow with the Institute of Multimedia Communication and Signal Processing, Friedrich Alexander University of Erlangen-Nuremberg, Germany. He received the Award of National Excellent Ph.D. Thesis from the Ministry of Education of China in 2012. His research interests include multimedia signal processing and perceptual signal processing.
\end{IEEEbiography}

\begin{IEEEbiography}[{\includegraphics[width=1in,height=1.25in,clip,keepaspectratio]{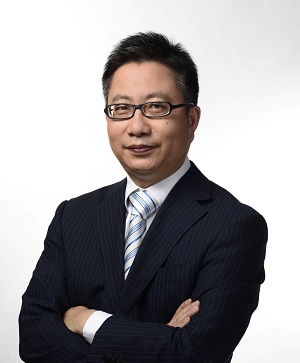}}]{Xiaokang Yang}(Fellow, IEEE)
received the B.S. degree from Xiamen University, Xiamen, China, in 1994, the M.S. degree from the Chinese Academy of Sciences, Shanghai, China, in 1997, and the Ph.D. degree from Shanghai Jiao Tong University, Shanghai, in 2000. From September 2000 to March 2002, he worked as a Research Fellow with the Centre for Signal Processing, Nanyang Technological University, Singapore. From April 2002 to October 2004, he was a Research Scientist with the
Institute for Infocomm Research (I2R), Singapore. From August 2007 to July 2008, he visited the Institute for Computer Science,
University of Freiburg, Germany, as an Alexander von Humboldt Research Fellow. He is currently a Distinguished Professor with the School of Electronic Information and Electrical Engineering, Shanghai Jiao Tong University. He has published over 200 refereed articles and has filed 60 patents. His current research interests include image processing and communication, computer vision, and machine learning. He is an Associate Editor of the IEEE TRANSACTIONS ON MULTIMEDIA.
\end{IEEEbiography}

\end{document}